%% file: thesis.tex
\documentclass[a4paper,openany]{cam-thesis}

\input{chapters/01Preamble}

\input{chapters/02ThesisInfo}
\input{chapters/WordCount}

\input{chapters/03Abstract}

\input{chapters/04Acknowledgments}

\begin{document}

\frontmatter

\ContentTitleFormat

\input{chapters/11Introduction}

\input{chapters/12Background}

\input{chapters/13MainChapter}

\input{chapters/15ResultsEvaluation}

\input{chapters/18Conclusion}

\renewcommand{\bibname}{References}
\newpage
\phantomsection
\addcontentsline{toc}{chapter}{References}
\bibliographystyle{plainnat}
\bibliography{Bibliography}

\input{chapters/21Appendix}

\end{document}

%% file: chapters/02ThesisInfo.tex
\title{Exploring Bad Minima in Transfer Learning}

\author{Mina Remeli}

\candidatenum{2118L}

\college{Churchill College}

\collegeshield{CollegeShields/Churchill}

\submissiondate{June, 2022}

\date{6 June 2022}

\subjectline{Computer Science}
\keywords{transfer learning, data poisoning, adversarial machine learning}

%% file: chapters/WordCount.tex
\wordcount{

Total page count: 57

Main chapters (excluding front-matter, references and appendix):
35~pages
(pp~7--41)

Main chapters word count: %
  \immediate\write18{texcount -1 -sum -merge -q thesis.tex output.bbl > thesis-words.sum }%
  \input{thesis-words.sum} words%

Methodology used to generate that word count:

\texttt{texcount -1 -sum -merge -q \#1.tex output.bbl \textgreater{} \#1-words.sum}

}

%% file: chapters/03Abstract.tex
\abstract{%
Transfer learning is a highly successful learning paradigm in deep learning which leverages knowledge from one task and applies it to another.
Despite its success, the necessary conditions for knowledge transfer are unclear and require further research.
In this dissertation, our aim is to understand transferability a little better by answering the following question:
Does good generalization on a source task necessarily imply transferability to a related target task?
We hypothesize the answer is negative, and attempt to find counterexamples via adversarial attacks on the transfer learning procedure.
Over three types of adversarial attacks, our empirical results show a strong relationship between generalization on the source task and transferability. 
These results do not rule out generalization on the source task as a sufficient condition for successful transfer.
}

%% file: chapters/04Acknowledgments.tex
\acknowledgements{%
  I would like to express my sincere gratitude to my supervisor, Ferenc Huszár for his guidance on the direction of this research and for his supportive feedback. I would also like to thank Nitarshan Rajkumar for all the helpful discussions and constant support.
}

%% file: chapters/11Introduction.tex
\chapter{Introduction}
\label{firstcontentpage}
In recent years, deep learning has emerged as a highly successful learning paradigm for various tasks such as computer vision and natural language processing. Initially, most of the success of deep learning came from scaling the model's parameters and the size of its dataset. Considerable effort went into curating large labeled datasets such as ImageNet \cite{deng_imagenet_2009} - the successor of its natural language equivalent, WordNet \cite{miller_wordnet_1995} - which consists of more than 14 million annotated images. However, recent trends have begun researching methods that work well even when there are few, or even no labeled samples available. Some notable examples are few-shot learning \cite{wang_generalizing_2020}, zero-shot learning \cite{wang_survey_2019}, transfer learning \cite{pan_survey_2010} and self-supervised learning \cite{jing_self-supervised_2021}. In this dissertation we will try to answer why some of these methods work so well, with a special focus on transfer learning.

Transfer learning \cite{weiss_survey_2016, pan_survey_2010, tan_survey_2018,zhuang_comprehensive_2021} enables us to reuse models that were trained on larger datasets, and fine-tune them on tasks for which we have less data. The goal of such fine-tuning would be to "transfer" knowledge from the original, upstream task to a different, but connected downstream task. The idea behind the method is that the feature representations obtained in the pretraining phase could be successfully re-purposed later. However, the success of this method has not been limited to cases where source and target domains are connected: indeed, it is standard practice in medical image classification to finetune models pretrained on ImageNet \cite{taher_systematic_2021}. 
Nevertheless, transfer learning can also result in lower test accuracy, which is called negative transfer \cite{zhang_survey_2021}. This phenomenon is most often attributed to the difference in distribution between source and target domain, but is also dependent on data quality and size \cite{wang_characterizing_2019}.

 So what are the conditions and contributing factors of success, or failure? What knowledge is being transferred, and what (if any) inductive biases aid transferability? Researching these questions are of utmost importance for both understanding and improving the efficiency and reliability of methods. 
 In this dissertation we chose to research the relationship between generalization and transferability in deep learning, in situations where positive transfer is possible. More specifically, we aim to investigate the question of whether good generalization on the source task is a \textit{sufficient} condition for transferability of the learnt representation. Our initial hypothesis is that the statement does not hold and our aim is to prove this through counterexamples: bad minima in the transfer setting which achieve good generalisation on the source task, but which do not transfer well to the target task. 
 
To this end, we designed adversarial attacks on the training process that bias it towards finding such bad minima.
This choice was heavily inspired by some of the more recent `generalization in deep learning' literature, where they use bad minima to derive insights about how and when models generalize \cite{zhang_understanding_2017,liu_bad_2021}.
To reach these bad minima they use a type of adversarial attack called label poisoning.
 Adversarial attacks have already played a prominent role in empirical studies on transfer learning. For example, transfer learning has been shown to fail in unsupervised domain adaptation (UDA) when subjected to label poisoning attacks \cite{mehra_understanding_2021}, and  \citet{xu_adversarially_2022} found that adversarial robustness is neither a necessary nor a sufficient condition for transferability.
If we were successful in finding bad transfer minima, it would imply that transferability is a result of inductive biases in the deep learning training paradigm, rather than a direct consequence of generalization on the source task. Similarly to how studies on bad minima \cite{liu_bad_2021,grosse_security_2020, huang_understanding_2020} illustrate that SGD is biased towards minima that generalise well, bad transfer minima would demonstrate that it has a further preference for finding features that transfer well.
Our results could potentially help us understand inductive biases that are preserved and propagated through transfer learning naturally.

First, we begin Chapter \ref{chap:main} by laying the groundwork for our hypothesis and formalizing bad transfer minima. Then, we introduce three attacks on transfer learning that aim to achieve misclassification on the target task \textit{exclusively}: adversarial optimization \cite{huang_understanding_2020}, label poisoning, and poisoning w/ warmup. These attacks are then empirically tested in Chapter \ref{chap:eval} on ``traditional'' transfer learning, and on transfer learning with unsupervised pretraining (using self-supervised learning techniques). Background knowledge on adversarial attacks, transfer learning and self-supervised learning is presented in Chapter \ref{chap:background}. We use CIFAR-100 for swift experimentation, and divide it into two parts where the bigger half is used as the source domain, and the smaller as the target domain data. Our findings from the experiments can be summed up as follows: 

\begin{itemize}
    \item We demonstrate that these attacks indeed lead to poor performance on the target task (subset of CIFAR-100).
    \item One can recover from these attacks by training on clean data from the source domain (but not with clean data from the target domain).
    \item We observe a nearly linear relationship between test performance on the source and target tasks (Fig. \ref{fig:all_attacks}).
    \item All attacks also result in decreased generalisation performance on the source task. \textbf{We did not find bad transfer minima that maintain generalisation on source while hurting transfer performance only.}
\end{itemize}

We conclude our findings in Chapter \ref{chap:conclusion} highlighting potential limitations and avenues for future exploration.

%% file: chapters/12Background.tex
\chapter{Background}
\label{chap:background}

In Section \ref{sec:transfer} we introduce transfer learning and place it in the context of deep learning.
Then, Section \ref{sec:SSL} presents self-supervised learning as an unsupervised pretraining method in transfer learning. Furthermore, we provide details on a contrastive learning method called SwAV.
We conclude with Section \ref{sec:attacks_on_learning}, where we talk about poisoning attacks and the impact attacks had on deep learning generalization research.

\section{Transfer learning} 
\label{sec:transfer}

Transfer learning (TL) is a learning paradigm which leverages learned knowledge from one task, and applies it to another. Typically the source domain has a large number of labeled samples, as opposed to the target domain where labeled data is more scarce. Labels are often hard / expensive to procure because specific domain knowledge is required from the annotators (for example in the medical imaging domain). The acquisition of datapoints can be expensive as well (for example of rare natural events that are hard to simulate - such as solar maxima\footnote{Solar cycle: \url{https://www.swpc.noaa.gov/phenomena/sunspotssolar-cycle}}, which only occur every 11 years). As a solution to this problem, transfer learning has been successfully applied in multiple domains (such as medical imaging \cite{taher_systematic_2021, van_opbroek_transfer_2015}, bug prediction in software \cite{nam_heterogeneous_2018}, semantic segmentation of slums \cite{wurm_semantic_2019} and natural language processing \cite{ruder_transfer_2019} in general) - meaning that transfer learning has yielded higher test accuracy on the target domain's test data than training the same network from scratch.

\paragraph{Homogeneous vs. heterogeneous}
We distinguish between \textit{homogeneous} and \textit{heterogeneous} transfer learning \cite{weiss_survey_2016}. We will be mostly focusing on homogeneous transfer learning. Homogeneous transfer learning occurs when the input feature space of the source and target domain match. Following the definitions of \citet{weiss_survey_2016}, we denote domain $\mathcal{D}$ as the combination of the input feature space $\mathcal{X}$ and and the marginal probability distribution $P(X)$, where $X \in \mathcal{X}$ (eg. $\mathcal{D} = \{ \mathcal{X}, P(X) \}$). To illustrate these on an example, let us look at the domain of medical images. There, $\mathcal{X}$ is the space of all possible images, while $P(X)$ is the marginal distribution that represents medical images (such as CT scans, MRIs, etc.). While the input feature space matches, it is often the differences in marginal distribution that homogeneous transfer learning has to account for. Heterogeneous transfer learning on the other hand is when the input feature space of source and target domain do not match. An example would be genetic vs. mammographic data - both can be used for breast cancer prediction, but have very different features. Heterogeneous transfer learning mostly deals with how these different input spaces can be aligned.

\paragraph{Source vs. target task}
Another axis of differentiation is whether or not the source and target tasks match. Following the definitions of \citet{weiss_survey_2016}, by target task $\mathcal{T}$ we mean the set of labels $\mathcal{Y}$ and the conditional distribution $P(Y|X)$. Illustrating these concepts in the context of sentiment analysis, $\mathcal{Y}$ is the binary label set representing $\{ \textit{positive}, \textit{negative}\}$, while $P(Y|X)$ would be the label distribution conditional to the input texts $X$. Now if the input texts are customer reviews, then some words might be associated with different sentiments, depending on where the reviews were sourced from (this is an example of a conditional distribution shift called context feature bias \cite{day_survey_2017}). For example the word ``heavy'' would typically carry positive meaning in the context of jewellery reviews (indicating that it is of high quality) while the same word could have a negative meaning in the context of the portability of tech gadgets.

\paragraph{Domain adaptation}
For clarification purposes, we must also mention \textit{domain adaptation}: an overlapping, but distinct concept from transfer learning. While there is some ambiguity in how the field uses this terminology, we will be following \citet{pan_survey_2010}'s definition. It identifies domain adaptation as a learning concept where the source and target tasks match, however the domains do not. In other words, the marginal probability distributions $P(X)$ are different. We often talk about \textit{unsupervised} domain adaptation \cite{wilson_survey_2020}, where we have labeled data for the source domain, but no labels for the target domain. For more on domain adaptation, please refer to the following surveys: \cite{jiang_literature_2007,wang_deep_2018}.

\paragraph{TL categories}
\citet{pan_survey_2010} identify the following categories in homogeneous transfer learning: instance transfer, feature-representation transfer, parameter transfer and relational-knowledge transfer.
The idea behind instance transfer is that while using the entire dataset of the source domain would generally hurt the target domain's performance, there are some parts of it that could be useful. They generally include various instance selection and weighting approaches. On the other hand, feature representation transfer tries to find a common latent representation which enables transfer from one task to another. Parameter transfer aims to re-use parameters of the model trained on the source domain, to leverage priors that were learned in the source task. 
For more on relational-based transfer learning, please refer to \citet{pan_survey_2010, weiss_survey_2016}.

\paragraph{Deep transfer learning}
In recent years, the focus of transfer learning research has shifted: the mounting success of deep learning  produced a whole new wave of research on \textit{deep transfer learning} \cite{tan_survey_2018,wang_deep_2018,zhuang_comprehensive_2021}. In these latest surveys capturing this trend, some new categories were introduced which we will not elaborate on for the sake of brevity.
However, we do want to put parameter transfer into the context of deep learning, as nowadays that is the most widespread scenario when one talks about transfer learning in general\footnote{Based on google search results for `transfer learning'.} (and also the scenario most relevant to us).

Essentially, parameter transfer based methods re-use parts of a deep-learning network (trained on the source domain task), which are fine-tuned on the target domain. This fine-tuning can be performed on the entire network (including the re-used weights), or solely on the additional layers that enable the target task prediction. In deep transfer learning, the source task is often referred to as the \textit{upstream} task, while the target task is often referred to as the \textit{downstream} task. One of the first instances of such network reuse in the context of deep learning was done in \citet{oquab_learning_2014}, where CNN feature representations from one task were re-used in another. A similar approach was taken in \citet{huang_cross-language_2013}, where the feature transformation layers of the speech recognition model are shared between the target languages. These approaches result in an improved model performance over training from scratch on the target task - also known as positive transfer.

\section{Self-supervised learning}
\label{sec:SSL}
\paragraph{SSL and TL}
Self-supervised learning (or SSL) \cite{jing_self-supervised_2021} is a term for learning methods in the unsupervised setting (eg. where there is no labeled data) where some supervised signal is introduced typically through some \textit{pretext task} such as center-word-prediction \cite{mikolov_efficient_2013} in NLP or solving a jigsaw puzzle \cite{noroozi2016unsupervised} in computer vision. The emphasis is on the pretext (or auxiliary) task, which allows us to learn useful representations of our input data. These representations can then be applied to a wide variety of tasks by fine-tuning. Most often self-supervised methods' performance is tested in the context of transfer learning (by training a linear probe on the learned representations). While supervised pretraining methods can be prone to overfitting to one task and transfer poorly to others, self-supervised pretraining methods could offer representations that are more transferable (for example SSL outperforms supervised pretraining when the distribution shift between source and target domains is large \cite{yang_transfer_2020}).

\paragraph{SSL and few-shot / zero-shot learning}
Another domain where self-supervised training methods have performed remarkably is few-shot  \cite{wang_generalizing_2020} and zero-shot learning \cite{wang_survey_2019} where little to no data is needed to adapt a model to perform a different task.
This is especially true of large language models that can be prompted to perform a wide variety of tasks  such as translation, question answering and even reasoning - this process is called prompt engineering.
The extent of achievable tasks and engineering of such prompts has been greatly researched  \cite{liu_pre-train_2021}. 

\paragraph{SSL categories}
\citet{jing_self-supervised_2021} introduce four categories of pretext tasks in visual feature learning: generation-based, context-based, free semantic label-based and cross modal-based. Image colorization, image inpainting and super-resolution are typical examples of generation-based methods. They often work based on a generative or encoder-decoder architecture. Context-based methods are effective at learning the spatial context of an image through tasks such as the jigsaw puzzle \cite{noroozi2016unsupervised}, context \cite{doersch_unsupervised_2016} or geometric transformation prediction \cite{gidaris_unsupervised_2018}. Recently, contrastive learning (another context-based method) has been particularly successful: MoCo was the first unsupervised method to surpass supervised pre-training \cite{he_momentum_2020}. We will elaborate more on contrastive learning in Section \ref{sec:contrastive}. Continuing with the pretext task categories, free semantic label-based methods automatically generate labels from synthetic images where depth, contour and surface normal can be predicted \cite{ren_cross-domain_2017}. Last but not least, cross-modal based methods use multiple input channels to learn to match segments across different modalities \cite{korbar_cooperative_2018}.

\subsection{Contrastive learning} 
\label{sec:contrastive}
Contrastive learning \cite{jaiswal_survey_2021} is a context-based pretext task that has gained popularity in recent years due to its success. 
Contrary to a context-based method, or a generative task (such as image inpainting) it focuses on generating similar representations for similar inputs. Because no supervised information (eg. class labels) is available to define similarity between inputs, contrastive learning algorithms focus on individual samples instead. The general idea is to take an input $\mathbf{x}$ (often called the \textit{anchor}) and apply one (or several) transformations \footnote{These transformations do not change the semantics of the original image. They are typically image augmentations such as cropping, rotating, greyscaling, etc.} to produce \textit{positive} examples, whose representations should be similar to $\mathbf{x}$. Other images will be \textit{negative} examples. To achieve similarity of representation to positive examples $\mathbf{x}^+$, and dissimilarity from negative examples $\mathbf{x}^-$, contrastive learning literature uses the InfoNCE loss \cite{oord_representation_2019}, where $f(.,.)$ is some similarity measure:

\begin{equation}
    \mathcal{L}_\text{N}(\mathbf{x}, \mathbf{x}^+, \{\mathbf{x}^-_i\}^{N-1}_{i=1}) 
= -\log\frac{\exp(f(\mathbf{x}, \mathbf{x}^+))}{\exp(f(\mathbf{x}, \mathbf{x}^+)) + \sum_{i=1}^{N-1} \exp(f(\mathbf{x}, \mathbf{x}_i^-))}
\end{equation}

During training this effectively means that inputs are passed through an encoder network. The obtained feature representation of the augmented anchor will be the positive sample, while the rest of the samples in the mini-batch will act as the negative samples. Notice how learning to equate between an image and its augmented version is a pretext task in itself: instead of generating pseudo-labels predicting the exact geometric transformation that was applied to an image, it learns invariance to geometric transformations more organically.

\citet{jaiswal_survey_2021} categorize contrastive learning architectures into four distinctive categories based on how they procure their negative samples:  end-to-end, memory bank, momentum encoder and clustering. In end-to-end architectures there are two separate encoders for positive and negative samples. One of the most notable end-to-end approaches is SimCLR \cite{chen_simple_2020}. Better performance was often associated with more negative samples (bigger batch sizes). Therefore methods such as PIRL \cite{misra_self-supervised_2019} use memory banks to store and retrieve previously calculated feature representations. However, these can get outdated after a few iterations, and updating them is computationally expensive. MoCo \cite{he_momentum_2020} uses a momentum encoder instead to address these issues. Last but not least clustering approaches have shown very promising results recently: SwAV \cite{caron_unsupervised_2021} is less than 1\% behind in terms of accuracy compared to the fully supervised setting. They have extended positive samples from the anchor-augmentation pair to images belonging to the same cluster.

\subsubsection{SwAV}
SwAV takes inspiration from previous clustering-based SSL approaches (such as DeepCluster \cite{caron_deep_2019}), and integrates them into contrastive learning. Clustering-based methods alternate between two phases: a clustering and a learning phase. First, learned feature representations are assigned a cluster, then in the learning phase cluster assignments are predicted from feature representations. SwAV performs these two steps for each mini-batch, and in the process learns the optimal vectors representing the clusters, and the cluster assignments. We denote the learned \textbf{prototype vectors} as $\{\mathbf{c}_1, ..., \mathbf{c}_K\}$ that form matrix $\mathbf{C}$, and $\mathbf{Q}=[\mathbf{q}_i,...,\mathbf{q}_B]$ as the \textbf{codes}, which are the mapping of batch samples to prototype vectors. Let us denote the input image as $\mathbf{x}$, and its augmented view as $\mathbf{x}_{t}$ (where $t \sim \mathcal{T}$, where $\mathcal{T}$ is a set of image transformations). This image is passed through an encoder $f_{\theta}$, and mapped to a unit sphere, yielding us $\mathbf{z}_{t}$. To make sure that code assignments are consistent between different views of the same image (say $\mathbf{z}_{t}$ and $\mathbf{z}_{s}$, with corresponding codes $\mathbf{q}_{t}$ and $\mathbf{q}_{s}$), the authors use a ``swapped'' prediction problem, where $l(\mathbf{z}, \mathbf{q})$ is the cross-entropy loss.

\begin{equation}
    L(\mathbf{z}_t, \mathbf{z}_s) = l(\mathbf{z}_t, \mathbf{q}_s) + l(\mathbf{z}_s, \mathbf{q}_t)
\end{equation}

This means that the obtained representations $\mathbf{z}$ should be indicative of the assigned code, and $\mathbf{q}$'s should match as closely as possible for transformations of the same image. The softmax $\mathbf{p}_t^{(k)}$ inside the cross-entropy loss is computed between $\mathbf{z}_{t}$ and prototype vector $\mathbf{c}_k$, where $\tau$ is a temperature parameter.

\begin{equation}
    l(\mathbf{z}_t, \mathbf{q}_s) = - \sum_{k} \mathbf{q}_s^{(k)} \log \mathbf{p}_t^{(k)} \text{, where } \mathbf{p}_t^{(k)} = \frac{\exp{\frac{1}{\tau} \mathbf{z}_t^{\top} \mathbf{c}_k}}{\sum_{k'} \exp{\frac{1}{\tau} \mathbf{z}_t^{\top} \mathbf{c}_{k'}}}
\end{equation}

Taking this loss over all images (and their augmentations) in a batch, we obtain the SwAV loss function, which is optimized with respect to model parameters $\theta$ and prototypes $\mathbf{C}$. Codes $\mathbf{Q}$ are computed beforehand, in such a way that they maximise similarity between features and prototypes, while avoiding the trivial solution where all of the images have the same code (by way of the entropy term $H$) and avoiding uniform codes across prototypes by setting $\epsilon$ to a low value.

\begin{equation}
    \max_{\mathbf{Q} \in \mathcal{Q}} \text{Tr}(\mathbf{Q}^\top \mathbf{C}^\top \mathbf{Z}) + \epsilon H(\mathbf{Q})
\end{equation}

\section{Attacks on learning}
\label{sec:attacks_on_learning}
In the adversarial machine learning literature, there are two broad categories of attacks \cite{liu_privacy_2021}: adversarial and poisoning attacks. Adversarial attacks are imperceptible perturbations in the input that achieve misclassification and were first introduced by \citet{szegedy_intriguing_2014}. They are typically employed in the test phase. Poisoning attacks on the other hand inject corrupted samples in the training phase. The goal of the adversary can be to achieve misclassification in general (on all samples) \cite{yang_generative_2017}, it can be targeted \cite{shafahi_poison_2018}, or it can evoke some desired behaviour (backdoor attacks) \cite{chen_targeted_2017}. One does not even necessarily need to use corrupted samples - backdoor attacks can also be achieved via data reordering attacks \cite{shumailov_manipulating_2021}.

\subsection{Poisoning attacks}
One of the first to evaluate poisoning attacks in the context of Machine Learning (more specifically SVMs) was \citet{biggio_poisoning_2013}. Poisoning can be achieved either through changing the true label of a sample, or by watermarking, where the image is perturbed directly. Research on poisoning in the context of deep learning appeared only later, characterizing it as a bi-level optimization problem \cite{munoz-gonzalez_towards_2017}. On one hand, the adversary optimizes to maximize its adversarial goal, while simultaneously minimizing the original training objective. Solving this bi-level optimization problem is intractable in the context of deep learning, therefore approximate approaches have been proposed. In case of targeted poisoning, \textit{feature collision} approaches \cite{shafahi_poison_2018} are popular for their efficiency: they aim to bring the feature representations of the targeted image closer to an image of a different class. However, they mostly work in fine-tuning scenarios, where only a fraction of the architecture is fine-tuned on the poisoned data. More promising approaches include first-order optimization approaches such as MetaPoison \cite{huang_metapoison_2020} or gradient matching approaches \cite{geiping_witches_2021} that also work in scenarios where the models are trained from scratch. While there is an undoubtedly rich literature on targeted attacks, our focus will be on attacks that aim to achieve misclassification on all samples - or otherwise known as \textit{indiscriminate} attacks \cite{munoz-gonzalez_towards_2017}.

As mentioned before, the goal of indiscriminate poisoning attacks is to cause misclassification on \textit{any} sample, not just one. Again, since the bi-level optimization problem introduced by \citet{munoz-gonzalez_towards_2017} is intractable in case of deep learning, other approaches were explored. \citet{yang_generative_2017} and \citet{munoz-gonzalez_poisoning_2019} both propose GAN-based approaches, where the generator outputs poisoned samples. \citet{munoz-gonzalez_poisoning_2019} propose pGAN, that generates samples that maximize misclassification, while remaining similar to unpoisoned samples (the latter is enforced by the discriminator).

\subsection{Deep learning through the lens of attacks}
\label{sec:dl_attacks_generalization}
Attacks have also proven to be a useful tool in understanding generalization in deep learning.
However, this was not always the case.
In the `pre-DL era', generalization was understood from the statistical learning theory point of view.
An important aspect of it was finding upper bounds on the generalization error. 
This upper bound was often dependent on some measure of complexity - complexity which could be expressed in the number of training samples, number of parameters, etc.
From this point of view, deep learning with its millions of parameters seemed like an unlikely candidate for models that generalize well.
For more on statistical learning theory, please refer to \citet{vapnik_nature_2000}.
\citet{zhang_understanding_2017}'s work was key in shifting the research community's views. They have shown that overparameterized neural networks - despite their capacity to perfectly memorize random labels\footnote{They trained models on random labels. Their expectation was that since there was ``nothing to learn'', the model would not converge.} - have an inductive bias \cite{neyshabur_search_2015} towards solutions that generalize well.
Their results questioned whether statistical learning theory was indeed a good fit for explaining generalization in deep learning.
And thus a shift towards researching inductive biases had begun.

One of the pivotal works that followed was by \citet{liu_bad_2021}, which used adversarial weight initializations to demonstrate the important role regularization (as an explicit bias) plays in generalization. First, they overfit the model on randomized labels (which led to the adversarial weight initialization), then they proceeded with normal training (without any regularization). They observed a drop in test accuracy of up to 40\%, compared to models that were randomly initialized. However, if regularization was re-introduced, the models escaped those bad minima. 
They demonstrated that there exist cases (bad initializations) where the implicit bias of SGD is not enough, and thus generalization in deep learning can be compromised.
Even something as simple as changing the scale of the initialization \cite{mehta_extreme_2021} or permuting the weights \cite{grosse_security_2020} can induce this kind of behaviour. Bad minima can also be reached by changing the loss function that we use during optimization \cite{huang_understanding_2020}. While evidence suggests that ``bad minima are everywhere'' \cite{huang_understanding_2020}, SGD often manages to find highly generalizing solutions.
Similarly to the approaches that are presented here, we too believe that researching the ``failure modes'' of deep learning is an important aspect of understanding the conditions and guarantees of generalization.

%% file: chapters/13MainChapter.tex
\chapter{Attacking Knowledge Transfer}
\label{chap:main}
We begin this chapter by motivating our work in Section \ref{sec:motivation}, where we give some insight into what is currently known about why transfer learning works.
Then, we introduce our attack-based approach for our research.
In Section \ref{sec:methodology} we define our central hypothesis, and outline the methodology with which we conduct our research. 
We follow that by a definition on bad transfer minima in Section \ref{sec:bad_transfer} and our proposed attacks in Section \ref{sec:attacks}. 
We conclude this chapter with details on our experimental setup in Section \ref{sec:exp_setup}.

\section{Motivation} 
\label{sec:motivation}
Transfer learning is currently one of the most successful training paradigms - it is widely used in domains such as medical imaging \cite{taher_systematic_2021, van_opbroek_transfer_2015} to natural language processing \cite{ruder_transfer_2019} where data on the target domain is typically scarce.
However, it is not guaranteed to work - sometimes transfer learning can produce a worse outcome than directly training from scratch on the target domain.
It is no wonder therefore that much of the research in this space is dedicated to understanding when, how and why transfer learning works best.
It is our goal too, to further this line of research.
First, we begin with a quick overview of what is currently understood about when transfer learning performs best.
Then, we proceed onto introducing our approach, and what we hope to contribute with our work.

\subsection{Related work on transferability}

The literature on transferability can be roughly divided into two broad categories: one is focused on when transfer learning fails (otherwise known as negative transfer) and the other when it succeeds. The latter is mainly concerned about what conditions are best to facilitate transfer. What type of pretraining to use, choice of loss function, best finetuning strategy, etc. - basically anything that will increase test accuracy on the target domain. First we will share some results on positive transfer. Then, we will talk about negative transfer, and the main identified reasons behind transfer failure.

\citet{kornblith_why_nodate} evaluate 9 different objective functions in the transfer learning setting, and conclude that loss functions which produce better accuracy on the source domain do not result in better transfer. A similar conclusion was made by \citet{abnar_exploring_2021}: they explore the limits of large-scale pretraining, and conclude that accuracy increase on the source domain does not necessarily mean the same on the target task (it saturates). Furthermore, accuracy that is too high can even \textit{hurt} target task performance - this is also observed by \citet{liu_towards_2019}. They also highlight the importance of the similarity between source and target task labels. \citet{mensink_factors_2021} conduct a more large-scale practical analysis as to when transfer learning works best: for most target tasks they found that there exists a source domain which significantly outperforms ImageNet (which is often used as a `de-facto' source domain in pretraining).
As to exactly \textit{what} is being transferred, \citet{neyshabur_what_2021} show that there is a big emphasis on the reuse of learned features.

Unfortunately, positive transfer does not always occur. In some cases, applying transfer learning can result in lower test performance compared to training on the downstream task from scratch. This phenomenon is called \textit{negative transfer} \cite{rosenstein_transfer_2005}, and it is of critical importance in the transfer learning research space. Most often it is attributed to the source and target domain being too different, however other factors may be at play as well: the choice of transfer method, source and target data quantity and quality \cite{zhang_survey_2021,wang_characterizing_2019}. \citet{zhang_survey_2021} conclude that much of the research was focused on measuring domain similarity \cite{bernico_investigating_2018,meiseles_source_2020} or various mitigation strategies \cite{wang_transferable_2020,wang_characterizing_2019}.

\subsection{Attacking knowledge transfer}
In Section \ref{sec:dl_attacks_generalization} we already spoke about the significance of studying ``failure modes'' in deep learning, and how it can lead to useful insights about generalization.
Here, we wish to apply the same principle to study transferability - through the lens of attacks.
Adversarial attacks have already played a significant role in studying transfer learning.
For example, transfer learning has been shown to fail in unsupervised domain adaptation (UDA) when subjected to label poisoning attacks \cite{mehra_understanding_2021}, and  \citet{xu_adversarially_2022} found that adversarial robustness is neither a necessary nor a sufficient condition for transferability.

In this dissertation, we chose to study the relationship between generalization on the source task and transferability, where positive transfer is possible. 
More specifically, we aim to investigate the question of whether good generalization on the source task is a \textit{sufficient} condition for transferability of the learnt representation. Our initial hypothesis is that the statement does not hold and our aim is to prove this through counterexamples: bad minima in the transfer setting which achieve good generalisation on the source task, but which do not transfer well to the target task. 
Essentially, we aim to find attacks that would ``block'' knowledge transfer between the two tasks.

Our work is highly motivated by \citet{zhang_understanding_2017}'s contributions, that ultimately led to a shift in focus from statistical learning theory to understanding inductive biases of SGD (see Section \ref{sec:dl_attacks_generalization}).
Similarly, our work might shift focus \textit{away} from studying transferability solely as a byproduct of similarity between domains and good performance on the source task.
If we were successful in finding bad transfer minima, it would imply that transferability is a result of inductive biases in the deep learning training paradigm, rather than a direct consequence of generalization on the source task. Similarly to how studies on bad minima \cite{liu_bad_2021,grosse_security_2020, huang_understanding_2020} illustrate that SGD is biased towards minima that generalise well, bad transfer minima would demonstrate that it has a further preference for finding features that transfer well.

\section{Methodology}
\label{sec:methodology}

As outlined in the previous section, it is still unclear what (if any) role implicit biases play in positive transfer. 
To unveil this, we propose to study the connection between generalization and transferability.
We know that generalization on the source task is not a \textit{necessary} condition, since there are examples of models that generalize perfectly on the source domain, but result in negative transfer on the target domain - possibly due to the two domains being too different. But how about the case where there is no domain discrepancy\footnote{And no other factor which could potentially result in negative transfer - such as insufficient / poor quality data in the source domain.}? From here on out, let us assume that our source and target domain are sufficiently similar such that positive transfer is possible. Is generalization on the source task a \textit{sufficient} condition then? In other words, does there exist a scenario where the pretrained model generalizes well on the source domain, but does not transfer well onto the target domain - or rather, is generalization on the source domain enough for positive transfer? We begin by introducing our own hypothesis:

\begin{hyp}[Central hypothesis] \label{hyp:first}
In cases where positive transfer should be possible, there exists such a minimum which generalizes well on the source task, but does not transfer to the target task (\textit{bad transfer minimum}).
\end{hyp}

To verify the first part of the hypothesis (``transfer should be possible''), it is sufficient to demonstrate positive transfer between source and target task.
Meaning that first we should perform transfer learning between source and target task, and demonstrate that it is successful.
The second half of the hypothesis presumes that this does not always hold. We believe that there should exist a \textit{bad transfer minimum} in overparameterized neural networks which is a good solution for the source task, but transfers poorly onto the target task. If we can prove our hypothesis to be true, that would mean that the inductive biases behind transfer learning come from the training process itself, and are not a direct consequence of generalization on the source task.
Answering this question could potentially bring us closer to understanding the reason behind the success of transfer learning. 

To address this question, we first provide a formal definition for bad transfer minima.
Then, we test our hypothesis empirically, by designing various attacks that aim to them. We evaluate our attacks on the widely popular transfer learning scenario, where we pretrain in a supervised manner, as well as a self-supervised learning scenario, where training on the source domain is unsupervised.

\section{Bad transfer minima}
\label{sec:bad_transfer}
\paragraph{Notation}
Let $\mathcal{S}$ be the source task and $\mathcal{T}$ be the target task, with relevant test error measures $\epsilon_{\mathcal{S}}$ and $\epsilon_{\mathcal{T}}$, respectively. Let $f_{\mathcal{T}}$ denote the transfer/fine-tuning process that takes a pre-trained parameter $\theta_\mathcal{S}$ and returns updated parameters $\theta_{\mathcal{S}\rightarrow\mathcal{T}} = f_\mathcal{T}(\theta_\mathcal{S})$ which have been fine-tuned for task $\mathcal{T}$.

The goal of this dissertation is to find bad transfer minima, which we define relative to a baseline ``good'' minimum as follows:

\begin{definition}[Bad transfer minimum] \label{def:bad_transfer}
 Let $\theta_{\mathcal{S}}$ be the parameter obtained on task $\mathcal{S}$ in a baseline transfer scenario before fine-tuning for task $\mathcal{T}$. We define $\theta^\ast$ to be a bad transfer minimum relative to $\theta_{\mathcal{S}}$ if it satisfies the following two conditions:
\begin{enumerate}
    \item The change in error on the source task $\mathcal{S}$ is sufficiently small: \\ $\epsilon_{\mathcal{S}}(\theta^*) - \epsilon_{\mathcal{S}}(\theta_{\mathcal{S}}) < \Delta_{\mathcal{S}}$,
    
    \item Transfer error is sufficiently higher compared to baseline: \\ $\epsilon_{\mathcal{T}}(f_\mathcal{T}(\theta^*)) - \epsilon_{\mathcal{T}}(f_\mathcal{T}(\theta_\mathcal{S})) > \Delta_\mathcal{T}$,
\end{enumerate}
for some constants $\Delta_\mathcal{S}$ and $\Delta_\mathcal{T}$.

\end{definition}

The constants $\Delta_\mathcal{S}$ and $\Delta_\mathcal{T}$ for which bad transfer minima exist are dependent on the problem at hand. In essence, we will be interested in mapping the Pareto-front for a transfer situation: for a given $\Delta_\mathcal{S}$, what is the largest $\Delta_\mathcal{T}$ we can find. In our experiments on CIFAR-100, $\epsilon_{\mathcal{S}}$ and 
$\epsilon_{\mathcal{T}}$ denote the the $0-1$ test loss ($100\%$ - test accuracy), the fine tuning process $f_\mathcal{T}$ is linear probe transfer (fine-tuning the last fully connected layer).
To put it more plainly, the first condition requires us to find a minimum which is nearly as good as the baseline on the source task.
Agreeing to an appropriate value for $\Delta_\mathcal{S}$ is relative, and is thus left to be judged on a case-by-case basis.
The second condition demands the increase in error on the target task to be sufficiently large.
But what is a sufficiently large value for $\Delta_\mathcal{T}$?
For our purposes, we will set $\Delta_\mathcal{T}$ such that it describes when transfer learning fails - i.e. when negative transfer occurs.

\section{Attacks} 
\label{sec:attacks}
Next, we present three attacks (methods for finding bad transfer minima): adversarial optimization, poisoning, and poisoning with warmup. The goal of these attacks is to induce misclassification on the downstream data, in essence, by adversarially initializing parameters before the fine-tuning step. This is achieved by poisoning the pre-training process (either via modifying data, or the loss).

\begin{figure}
    \centering
    \includegraphics[width=\textwidth]{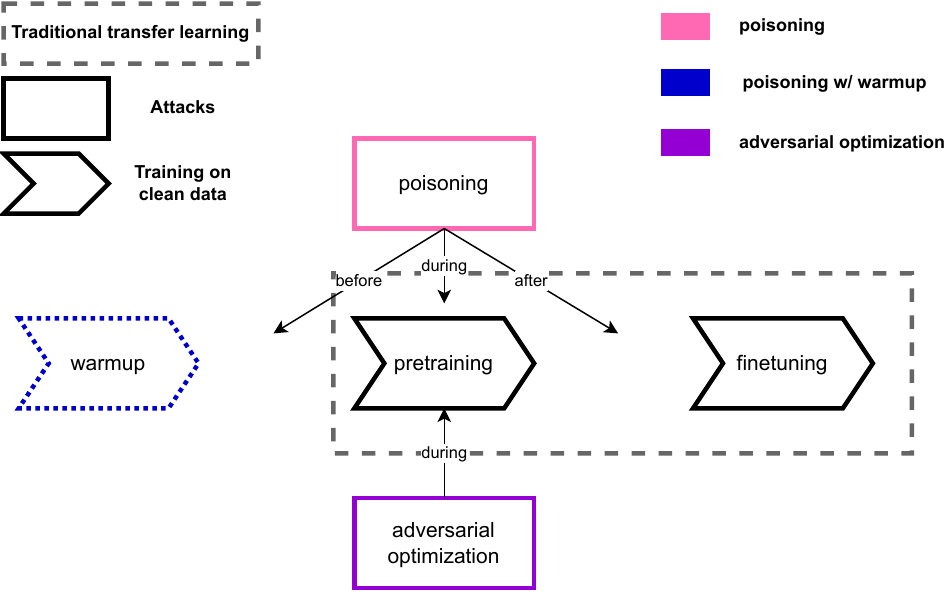}
    \caption{We introduce three attacks for achieving bad transfer minima: (1) adversarial optimization (2) poisoning and (3) poisoning w/ warmup. This figure illustrates when each attack occurs, relative to the different stages of transfer learning.}
    \label{fig:my_label}
\end{figure}

\subsection{Adversarial optimization}
Here we introduce the optimization-based attack from \citet{huang_understanding_2020}. They use a poisoned loss to lead the model to a poorly generalizing minimum. It is calculated on a mix of clean data ($D_t$) and poisoned data ($D_p$). Note that the poisoned data is not compromised in any sense, it just means that the cross entropy loss is calculated differently. Contrary to the clean data the input probabilities for the loss function are ``flipped'':

\begin{equation} \label{eq:poison_loss}
    L(\theta)=\frac{(1-\beta)}{|D_t|} \sum_{(x,y) \in D_t} - \log p_{\theta}(y|x) + \frac{\beta}{|D_p|} \sum_{(x,y) \in D_p} - \log{[1-p_{\theta}(y|x)]}
\end{equation}

, where $\beta$ is the poison fraction (fraction of poisoned data in a single minibatch). The first term (promoting regular training) in this equation will be the cross entropy loss calculated on $D_t$, while the second (promoting poisoned training) will be the \textit{reverse} cross entropy loss calculated on $D_p$. The second term of Eq. \ref{eq:poison_loss} will only be minimized if the input sample is misclassified.

\subsection{Label poisoning}
Next, we introduce a data poisoning attack, where the adversary randomizes the labels in order to decrease global testing accuracy. This type of poisoning attack is used in \citet{liu_bad_2021} \textit{before} training on the `clean' data in order to force the model into a bad minimum. In case of transfer learning, this poisoning can occur in three places: \textit{before}, \textit{during}  and \textit{after} pretraining. Poisoning before and after pretraining means that the adversary performs some number of iterations on the poisoned data exclusively. During pretraining the poisoned data is mixed with the clean data during updates. We call the ratio of poisoned samples to clean samples the \textit{poison ratio} $p$. 
There is some literature on poisoning attacks targeting unsupervised domain-adaptation \cite{mehra_understanding_2021}. In another study, pretraining has been found to be one of the most robust domain adaptation heuristics - even against label poisoning \cite{wiles_fine-grained_2021}.

\subsection{Poisoning with warmup}
The third kind of attack that we run experiments on is based on the label poisoning attack that we introduced in the previous section. The motivation behind it was to design an attack which only minimally affects the test accuracy on the source domain.
With this attack, the upstream pretraining is split into two stages: first, we perform some number of training iterations on clean upstream data, followed by training on a mix of poisoned and clean data (poisoning during pretraining). 
Inspired by the EWC loss from \citet{kirkpatrick_overcoming_2017} - which is used to prevent catastrophic forgetting - we initially experimented with adding an $l2$ anchor term to the loss and varying the learning rate during the poisoning, however those did not give us the desired results (for more on these experiments please refer to Appendix \ref{sec:A-anchor}).

\section{Experimental setup}
\label{sec:exp_setup}

\subsection{Data} \label{sec:data}
We use the CIFAR-100 dataset \cite{krizhevsky2009learning} for quick experimental results on transfer learning. It consists of 60000 32x32 images of 100 classes, with 500 train and 100 test samples per class. For the transfer learning setting, we use the first 90 classes for the pretraining phase (45000 train / 9000 test samples), and the last 10 classes for the finetuning phase (5000 train / 1000 test samples). From now on we refer to classes 0-89 as the \textbf{upstream task} and classes 90-99 as the \textbf{downstream task}. The downstream task labels are `train', `trout', `tulip', `turtle', `wardrobe', `whale', `willow tree', `wolf', `woman' and `worm', which represents a mix of various categories: it covers 10 out of the 20 superclasses that are defined for CIFAR-100.
Hence, there is no significant marginal distribution shift between the source and target domain, which should assist positive transfer.

\subsection{Models}
For the majority of the experiments on the CIFAR-100 dataset we use the ResNet-18 \cite{he_deep_2015} architecture (with 11.174M parameters), with 3x3 sized filter in the conv1 layer and no max pooling after conv1. This modification of the architecture is standard practice for the CIFAR-100 dataset, and is also used by \citet{liu_bad_2021}. We also use a ResNet-50 architecture (with 25.6M parameters) that was pre-trained on ImageNet \cite{deng_imagenet_2009} using a self-supervised learning method called SwAV\footnote{SwAV model zoo: \url{https://github.com/facebookresearch/vissl/blob/main/MODEL_ZOO.md\#SwAV}} \cite{caron_unsupervised_2021}.

\subsection{Training settings}

\subsubsection{Transfer learning}
We run experiments in a homogeneous transfer learning setting, which can be divided into two phases: pretraining and finetuning. In the pretraining phase we train on the upstream task in a \textit{supervised} manner, while in the finetuning task we re-use learned feature representations and train on the downstream task.
We evaluate all three attacks in this setup, and
we test two settings - one where we fine-tune only the last layer of the model (linear probe or \textit{LP transfer}) and where we fine-tune the entire model (\textit{full transfer}). Because the effect of the attacks was not as prominent in the latter case, we present the full transfer results in Appendix \ref{sec:A-full_transfer}.

\subsubsection{Self-supervised learning: SwAV}

SwAV \cite{caron_unsupervised_2021} is a new, highly successful self-supervised learning (SSL) method in the family of contrastive learning algorithms. 
We use SwAV  to test our attacks in a setting where the pretraining is \textit{unsupervised}. 
Initially, we wanted to make our experiments on SwAV as similar as possible to our supervised transfer learning experiments.
Originally our plan was to train a ResNet-18 model on CIFAR-100 data with SwAV.
However, we were not able to get state-of-the-art results for the baseline experiments, and running a hyperparameter sweep would have been too costly (for more details, please refer to Appendix \ref{sec:A-swav}).
Instead, we use ResNet-50 model pretrained on ImageNet \cite{deng_imagenet_2009}, and evaluate the `poison after pretrain` attack on it.

\subsection{Hyperparameters}
\paragraph{Transfer learning}
We use the same hyperparameters as in the 'vanilla SGD' setting of  \citet{liu_bad_2021} for a ResNet-18 architecture trained on CIFAR-100. That means an SGD optimizer without any regularization and a learning rate of 0.1. For batch size we chose 256 instead of 128, because we found that it resulted in more stable training (for more, please see Appendix \ref{sec:A-bs_128}). We also did not use any type of image augmentations for the training process. This was done so that we could isolate the effects of the attack, without any potential masking effect from regularization. Each stage of the training (pretraining, poisoning, finetuning) is 100 epochs long, unless indicated otherwise. Warmup is 20 epochs long \footnote{Based on the pretraining learning curves, we found that it converged after 20 epochs on CIFAR-100 upstream data. Please see Appendix \ref{sec:A-anchor}.}. The implementations for the experiments on transfer learning can be found in the following repository: \url{https://anonymous.4open.science/r/BadMinima-06D9}.

\paragraph{Self-supervised learning}
For the self-supervised learning experiments, we used the Vissl framework \footnote{Vissl: \url{https://github.com/facebookresearch/vissl}} for self-supervised learning research. 
Our base is a ResNet-50 model pretrained on ImageNet \cite{deng_imagenet_2009}, and finetune the entire model on poisoned downstream data for 100 epochs. We resize the input images to match the dimensions of the ImageNet images (which is 224x224), and train on them using SGD with $0$ weight decay and $0.9$ momentum with $0.1$ learning rate and batch size of 128 \footnote{Finetuning on the poisoned data took around 3 days to complete on a V100 GPU.}. The images were randomly flipped horizontally. We trained linear probes on the feature embeddings for evaluation purposes. Those were trained for 28 epochs with an SGD optimizer with weight decay of 0.0005 and a momentum of 0.9 with a batch size of 256 (image transformations were the same as before). For full code and more detailed configurations for running the experiments please refer to the following repository: \url{https://anonymous.4open.science/r/vissl-4447}.

\section{Summary}
In this chapter, we introduced our proposed line of research, namely finding \textbf{bad transfer minima}, which we formally define in Section \ref{sec:bad_transfer}. Our central hypothesis is that they exist, and motivated by prior research on bad minima in DL generalization \cite{zhang_understanding_2017,liu_bad_2021}, we propose to find these minima using attacks. 
\textbf{If we can prove their existence, that would imply that SGD does not only find minima that generalize well, but also prefers minima that \textit{transfer well}.}
We propose three attacks (Section \ref{sec:attacks}) that we evaluate on transfer learning with supervised pretraining and unsupervised pretraining in the next chapter. For details on the experimental setup, please refer to Section \ref{sec:exp_setup}.

%% file: chapters/15ResultsEvaluation.tex
\chapter{Results and evaluation}
\label{chap:eval}

In this chapter, we evaluate the attacks introduced in Section \ref{sec:attacks} on transfer learning. More specifically, we evaluate LP or linear probe transfer, where we only fine-tune the last layer of the model (results for full transfer can be found in Appendix \ref{sec:A-full_transfer}). For more on the experimental setup, please refer back Section \ref{sec:exp_setup}.

The results can be divided into two sections, based on the learning paradigm used during pretraining. In Section \ref{sec:sup_transfer} we have supervised, while in Section \ref{sec:unsup_transfer} we have unsupervised pretraining. Both sections start by introducing baselines, where we evaluate transfer learning when there is no attack. Later, we compare performance under attack to those baselines.  For a quick overview of all the experiments that we have done, please refer to Table \ref{tab:all_experiments}.

\begin{figure}
    \centering
    \begin{subfigure}{\textwidth}
        \includegraphics[width=0.49\textwidth]{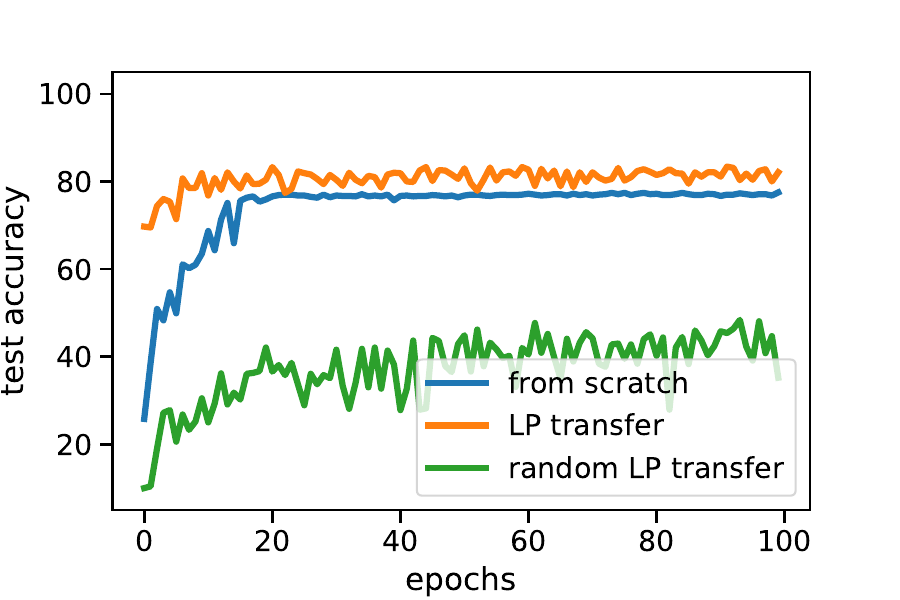}
    \includegraphics[width=0.49\textwidth]{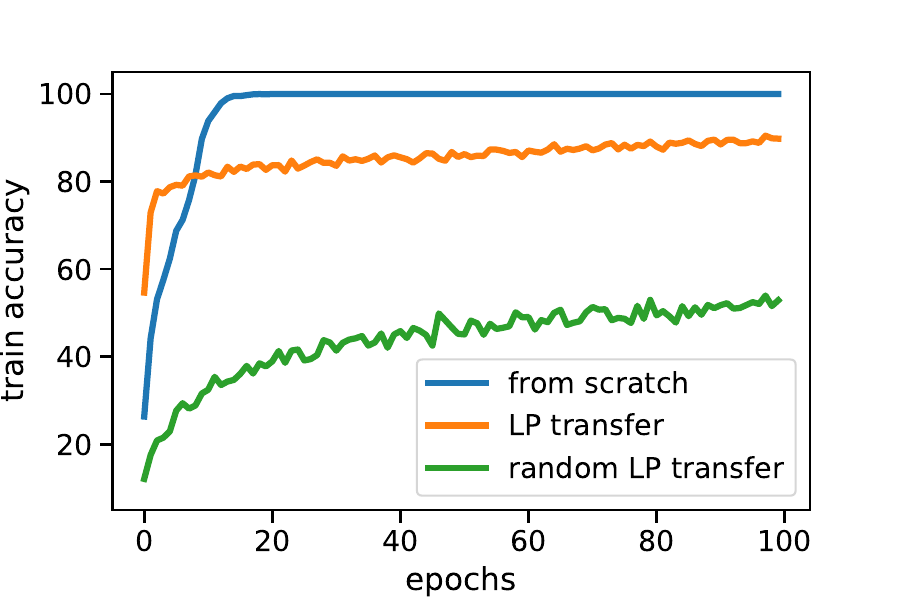}
    \caption{Learning curves of training on downstream data.  Plot is based on one run.}
    \label{fig:scratch_vs_transfer}
    \end{subfigure}
    
    \bigskip
    
    \begin{subfigure}{\textwidth}
        \begin{tabularx}{\linewidth}{lc|ccc}
    \toprule
              & \textbf{upstream} & \multicolumn{3}{c}{\textbf{downstream}}              \\
              &          & from scratch    & LP transfer & rand. LP transfer \\ \midrule
            test accuracy & $55.53\% \pm 0.62$ & $74.8\% \pm 0.36$ & $\textbf{80.97\%} \pm 0.91$ & $45.03\% \pm 0.31$ \\
    \bottomrule
    \end{tabularx}
    \caption{Achieved test accuracies on upstream and downstream data. Reported averages and standard deviations are based on 3 runs.}
    \label{tab:test_acc}
    \end{subfigure}

    \caption{We compare training from scratch, transfer learning (w/ linear probe) and transfer learning from a random initialized network. Reusing the feature representations from a pre-trained model generally result in a higher test accuracy, and faster convergence. %
    Generalization error (train - test accuracy) is also much lower in case of transfer learning. }
\end{figure}

\section{Transfer learning experiments}
\label{sec:sup_transfer}
\subsection{Baselines}

To have a better understanding of the success of an attack, first we must examine the model's performance \textit{without} it. In this section, we present downstream test accuracies when training from scratch and employing transfer learning, where we mean the following: first, the model is pre-trained on the upstream data, then we freeze the model weights except the last fully-connected layer which is fine-tuned on the downstream task (linear probe or \textit{LP transfer}). Furthermore, we also look at the naive worst-case transfer scenario: \textit{random LP transfer}, where we fine-tune on a randomly initialized network.

\subsubsection{Transfer learning is better than training from scratch}

Unsurprisingly, transfer learning improves generalization and results in faster convergence than training from scratch. On the CIFAR-100 downstream task, training from scratch leads to $74.8\%$ , while transfer learning with a linear probe can achieve $80.97\%$ average test accuracy ($\sim6\%$ increase).
While useful feature representations help training, random feature representations significantly lower the average test accuracy from $74.8\%$ to $45.03\%$ ( $\sim30\%$ decrease). 
On the upstream dataset, pretraining leads to $55.53\%$ average test accuracy.
For a summary of the reported accuracies please refer to Table \ref{tab:test_acc}.

\begin{figure}
     \centering
     \begin{subfigure}[b]{0.49\linewidth}
         \centering
         \includegraphics[width=0.87\textwidth]{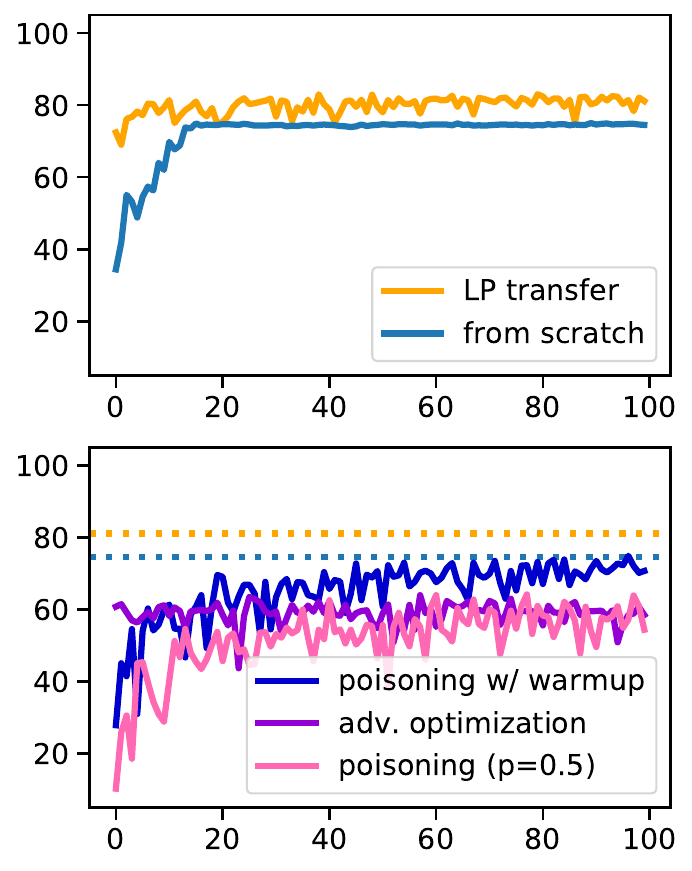}
         \caption{Finetuning phase learning curve (test accuracy).}
         \label{fig:learning_curves}
     \end{subfigure}
     \hfill
     \begin{subfigure}[b]{0.49\linewidth}
         \centering
         \includegraphics[width=\textwidth]{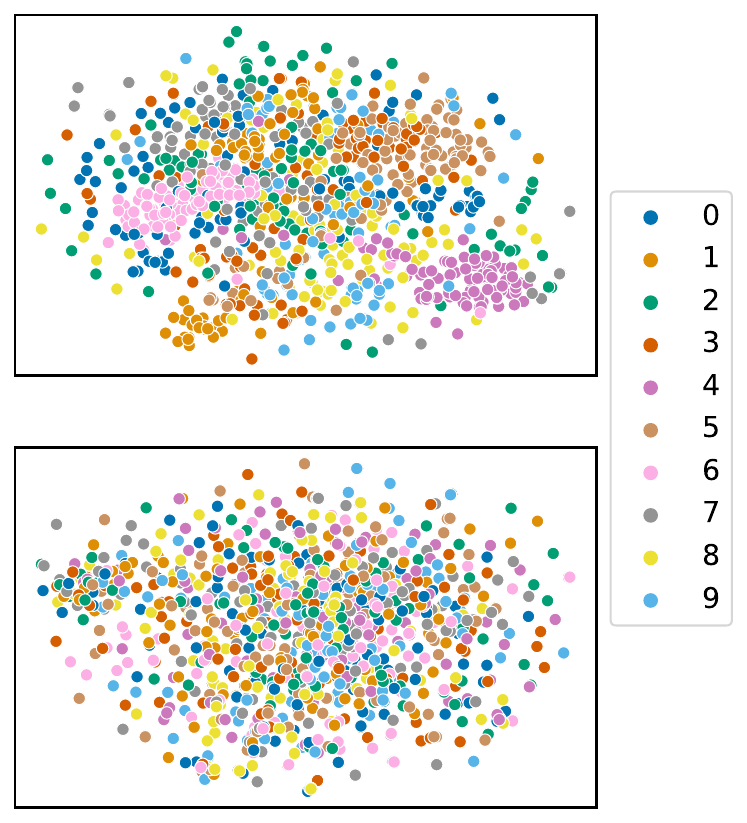}
         \caption{T-SNE of learned feature representations.}
         \label{fig:t-sne}
     \end{subfigure}
        \caption{Comparison of \textbf{no attack} (top figures) vs \textbf{attack} (bottom figures) scenarios. The learning curves in case of an attack become more jittery compared to models that were not attacked (it finds less robust solutions). Also the model seems to learn less useful feature representations (bottom feature representations were obtained with poisoning w/ warmup attack). Features are the output of the last residual block.}
        \label{fig:learning_curve_t-sne}
\end{figure}

\subsection{Attacks}

Next, we present our results on the attacks that were introduced in Section \ref{sec:attacks}: adversarial optimization, poisoning and poisoning with warmup. The goal of these attacks is to drive misclassification on the downstream data. Adversarial optimization does this with a modified cross-entropy loss function (see Eq. \ref{eq:poison_loss}), while poisoning based attacks train on random perturbed labels. In our transfer learning context, this means the following:

\paragraph{Adversarial optimization}
We use the loss from Equation \ref{eq:poison_loss} to drive misclassification on the downstream data in the pretraining phase. We run our experiments by sampling equal amounts of upstream data ($D_t$) and downstream data ($D_p$), thus setting $\beta=0.5$. Note that the downstream data is not modified in any way, only the loss term that we use for backpropagation is changed. This drives the model to not only perform well on the upstream data, but also misclassify on the downstream data at the same time. Pretraining with adversarial optimization also 100 epochs.

\paragraph{Label poisoning}
To achieve misclassification on the downstream data, we assign random labels to the downstream data in the 0-99 range. Poisoning can occur in three places: before, during and after pretraining. Poisoning before and after pretraining means that the adversary performs some number of iterations on the poisoned data exclusively. During pretraining the poisoned data is mixed with the clean data during updates, which is regulated by poison ratio $p$. In all three cases training on poisoned data lasts for 100 epochs unless otherwise indicated.

\paragraph{Poisoning w/ warmup}
In this scenario we precede poisoning during pretraining (with poison ratio $p=0.5$) with a warmup phase, meaning that we begin pretraining with $n$ epochs of training on `clean' upstream data. Again, poisoned pretraining lasts 100 epochs. The warmup phase is 20 epochs long.

\subsubsection{Attacks lead to bad minima on the downstream task}
\label{sec:bad_minima}
In general, we observe that all three attacks that were introduced in Section \ref{sec:attacks} lead to sub-optimal solutions on the downstream task. In Figure \ref{fig:learning_curves} we can see that the learning curves of the attacked models are more unstable (meaning that the test accuracy fluctuates more) as opposed to the ``clean'' transfer learning or training from scratch entirely. Such fluctuation in test accuracy could be a result of a learning rate that is too high. One argument against this is that this fluctuating behavior should then also be more prominent in the ``clean'' transfer learning case. An alternative explanation could be that the attacked model finds a bad minimum, which exhibits unstable behaviour whenever the parameters of the linear probe are changed a bit in each iteration. In addition to finding non-robust minima, unsurprisingly the attacks also result in lower test accuracy on the downstream data. 

We also look at the T-SNE projections of the feature representations for the test downstream data in Figure \ref{fig:t-sne}. This effectively means that we took the outputs of the last residual block (before the final average pooling and fully connected layer) and used T-SNE \cite{maaten_visualizing_2008} (SciKit implementation with default parameters\footnote{\url{https://scikit-learn.org/stable/modules/generated/sklearn.manifold.TSNE.html}}) to produce the 2-D embeddings. The figure shows more distinct clusters in case of the unattacked model. Of course the lack of structure in case of the attacked model's feature embeddings could be due to a poor choice of T-SNE parameters \footnote{How to Use t-SNE Effectively \url{https://bit.ly/39OgCIE}}. We did not do a hyperparameter search to validate this. 

\begin{figure}[h]
     \centering
     \begin{subfigure}[b]{0.49\textwidth}
         \centering
         \includegraphics[width=\textwidth]{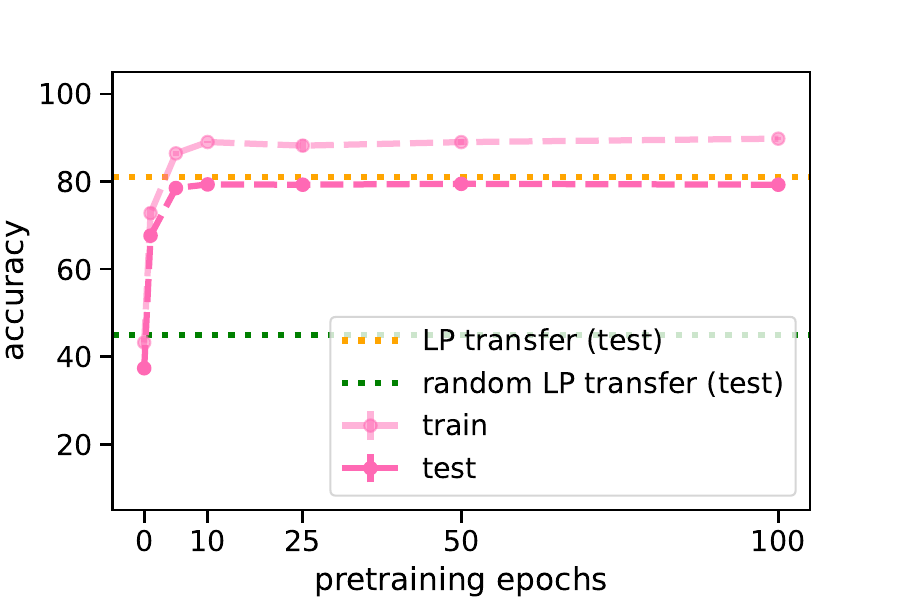}
         \caption{Poisoning before pretraining. Effect of pretraining length.}
         \label{fig:pretraining_length}
     \end{subfigure}
     \hfill
     \begin{subfigure}[b]{0.49\textwidth}
         \centering
         \includegraphics[width=\textwidth]{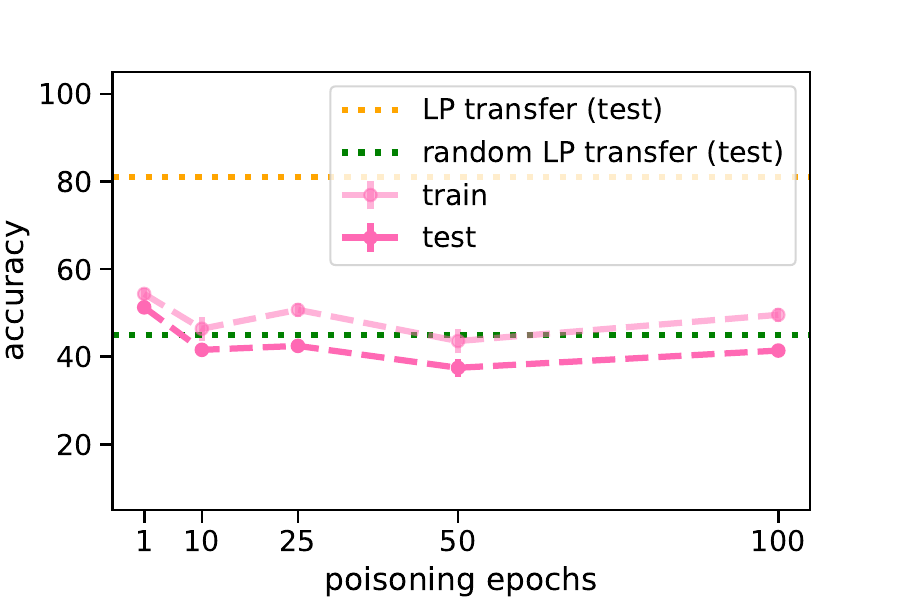}
         \caption{Poisoning after pretraining. Effect of number of poisoning epochs.}
         \label{fig:poison_len}
     \end{subfigure}
     \hfill
     \begin{subfigure}[b]{0.6\textwidth}
         \centering
         \includegraphics[width=\textwidth]{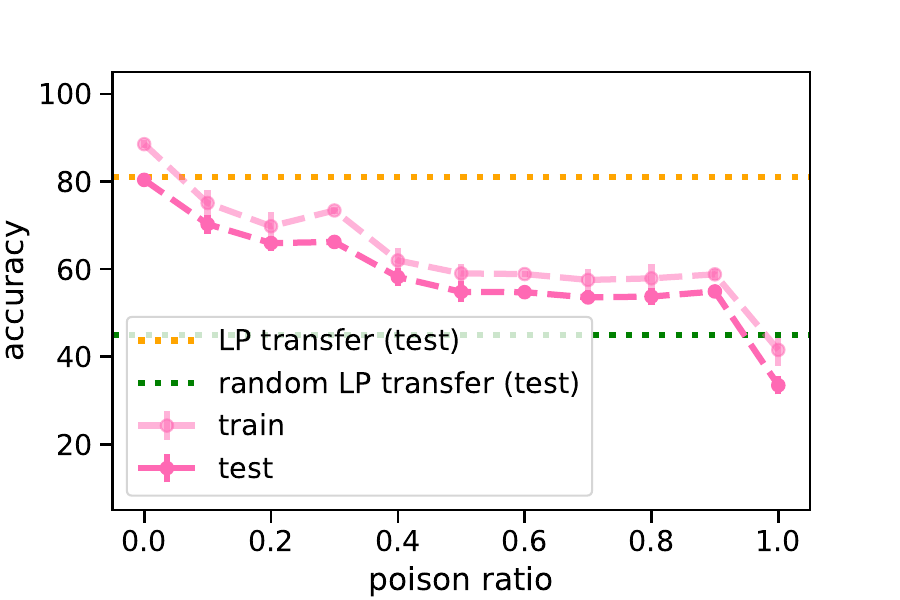}
         \caption{Poisoning during pretraining. Effect of poison ratio $p$.}
         \label{fig:poison_ratio}
     \end{subfigure}
        \caption{Effect of poisoning attack on finetuned downstream accuracy. We evaluate poisoning \textit{before}, \textit{during} and \textit{after} pretraining. We compare to the vanilla transfer learning setting (`LP transfer') and to transfer learning on a random initialized network (`random LP transfer'). Reported mean and SE is from 3 runs.}
        \label{fig:three graphs}
\end{figure}

Also, the effect of label poisoning is immediate: even one epoch of training on poisoned data is capable of massively deteriorating the downstream test accuracy (see Fig. \ref{fig:poison_len}). The poisoning can achieve an average $41.4\% \pm 1.51$ downstream test accuracy, which is a comparable downgrade in accuracy to the `random LP transfer' naive worst-case baseline ($45.03\%$). This is the scenario where poisoning occurs \textit{after} a pretraining phase on clean upstream data. It demonstrates how quickly the useful feature representations from the pretraining phase can be ``forgotten''. This is somewhat surprising, because one epoch is not even enough to learn the random labels (it achieves $2.38\%$ accuracy on the poisoned data). Another interesting observation is that after pretraining, accuracy on poisoned data does not reach $100\%$ - in fact, it plateaus at $34\%$ after 10 epochs. This suggests that pretraining could possibly condition the model not to memorize as much. 
It would be interesting to explore in further experiments whether models can indeed be conditioned not to memorize by pretraining on clean data.

The extent of the negative effect can also be varied. In our ``poison during pretraining'' setting we tried multiple values for  poison ratio $p$, to see how the amount of poisoned data in each minibatch affected downstream performance (see Fig. \ref{fig:poison_ratio}). In general, if we increase the amount of poisoned data in the pretraining phase, the finetuned downstream accuracy decreases accordingly. After $p > 0.5$, the finetuned downstream test accuracy does not decrease below $50\%$ (except for $p = 1.0$).

To summarize, \textbf{we found some evidence suggesting that attacked models find bad minima}:

\begin{itemize}
    \item We witnessed learning curves exhibiting unstable behaviour in Fig. \ref{fig:learning_curves}.
    \item Attacked models seem to learn less useful feature representations (Fig. \ref{fig:t-sne}).
\end{itemize}

We also present some findings as to how fast the model can switch to this undesired behaviour: 1 epoch of training on poisoned data is enough to lead to a bad minimum (Fig. \ref{fig:poison_len}). We also show that this negative effect can be scaled by using more poisoned data (Fig. \ref{fig:poison_ratio}).

 \begin{figure}
    \centering
    \includegraphics[width=0.55\textwidth]{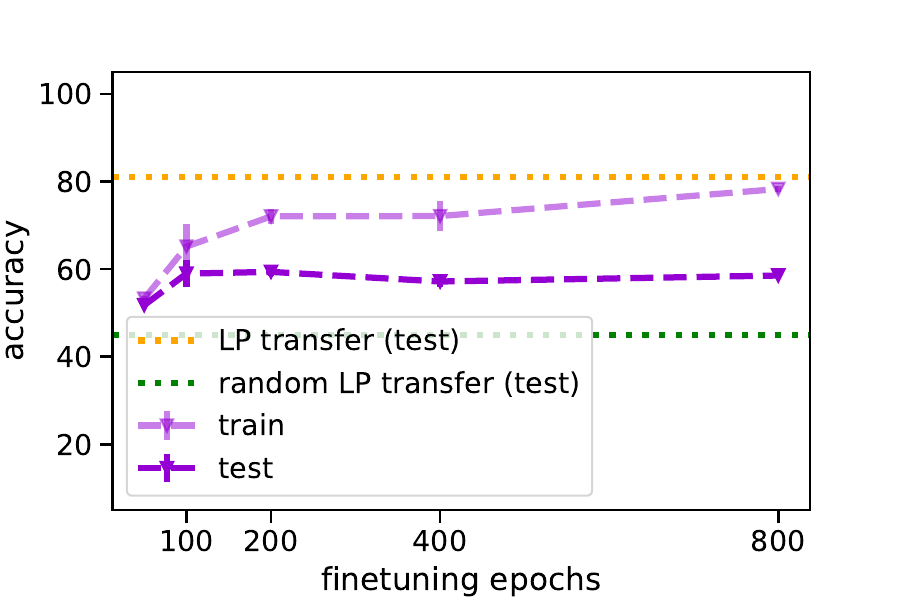}
    \caption{Downstream accuracies on network pretrained with adversarial optimization, then finetuned to downstream task. Increasing the number of finetuning epochs only increases the generalisation gap. We compare to the vanilla transfer learning setting (`LP transfer') and to transfer learning on a random initialized network (`random LP transfer'). Reported mean and SE is from 3 runs. 
    }
    \label{fig:adv_opt_finetune_len}
\end{figure}

\subsubsection{Clean training after attack: pretraining escapes bad minima, finetuning does not}
\label{sec:clean_training}

By now we have established that attacks lead to poorly generalizing minima on the downstream task. A relevant question to ask would be whether one can get out of such bad minima by training on clean data. If the effect of poisoning is reversible, that would give us a clue about how we can reach good solutions, even if the data/model has been compromised. After our model has been attacked, there are two options from which we can choose: we can either prolong the finetuning phase in hopes of our model recovering, or we can pretrain on upstream data that is 100\% clean (meaning it is high-quality data that does not contain poisoned samples). This can be a carefully curated subset of the upstream data.

 We run an experiment where we test varying finetuning lengths to see whether we can recover from poisoning (see Fig \ref{fig:adv_opt_finetune_len}). In the pretraining phase the model was trained using the adversarial optimization process, then it was finetuned on clean downstream data. Unfortunately, finetuning does not negate the effect of poisoning, even if we train for 8x longer than before. Longer periods of finetuning only seem to increase the training accuracy, increasing the generalization error and making the model overfit.

 On the other hand, if poisoning occurs before pretraining on clean data, then the effect of poisoning can be entirely reversed (see Fig. \ref{fig:pretraining_length})! First, we train the model on the poisoned data, then pretrain on the clean upstream dataset before finetuning on the clean downstream data. This effectively means that the attacker manipulates the model \textit{before} any of the transfer learning steps happen, which is the setting most similar to \citet{liu_bad_2021}. Before pretraining, train accuracy on poisoned data reaches $100\%$. After pretraining and finetuning however, the effect of poisoning entirely disappears. Even 10 epochs of pretraining on clean data is sufficient to recover the downstream test accuracy completely! 
 
 In general, the attacks clearly \textit{can} result in lower test accuracy compared to the `clean' transfer learning accuracy. If no pretraining is done at all, the effect of poisoning can be as bad as doing transfer learning on a random initialized network. And the effect of poisoning does not disappear by finetuning on clean downstream data for a longer period of time (Fig. \ref{fig:adv_opt_finetune_len}). However, \textbf{poisoning can be `undone' if we train on clean upstream data afterwards} (Fig. \ref{fig:pretraining_length}). This means that periodically `re-setting' the model by training on high-quality upstream data can help the model avoid non-robust solutions.

\subsubsection{Attacks also affect upstream task negatively}
\label{sec:us_neg_affected}
Our original goal was to find an attack which would condition the model to perform well on the upstream task, while performing poorly on the downstream task. Based on our definition in Section \ref{sec:bad_transfer} we expected to find an attack which satisfies the following: (1) it achieves nearly the same test accuracy on the upstream task (bounded by some $\Delta_\mathcal{S}$) and (2) achieves lower test accuracy on the downstream task compared to the baseline (negative transfer). Here, we set $\Delta_\mathcal{S}$ to be 5\%, which means that a 5\% degradation in upstream test accuracy is acceptable.

\begin{figure}
    \centering
    \includegraphics[width=0.8\textwidth]{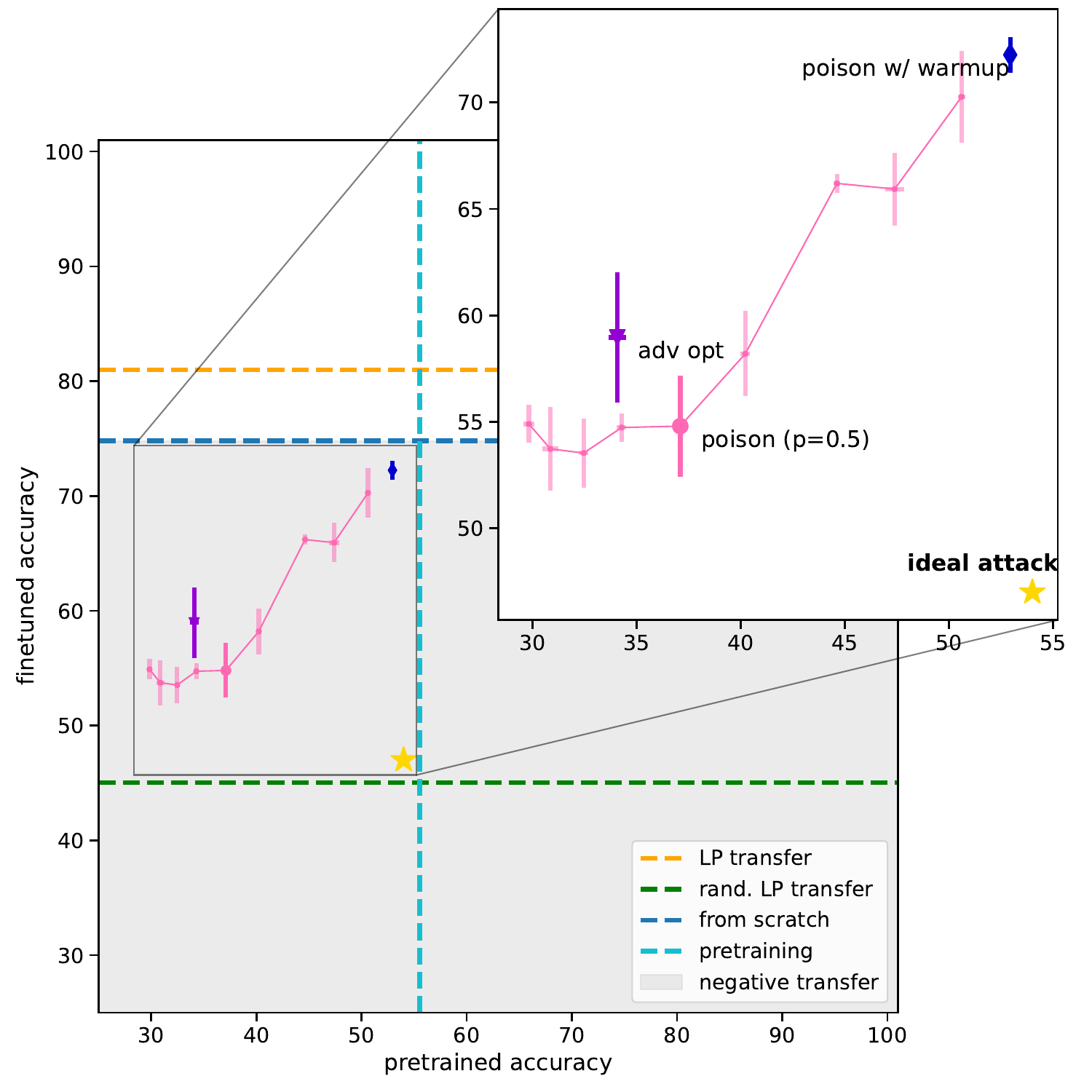}
    \caption{Test accuracies after the attacks. X axis is pretrained model accuracy on upstream data. Y axis is finetuned model accuracy on downstream data. Dashed lines are the baselines from Table \ref{tab:test_acc}.}
    \label{fig:all_attacks}
\end{figure}

While all of our attacks achieve negative transfer (all of them fall into the gray shaded area of Figure \ref{fig:all_attacks}), there seems to be a linear relationship between the upstream test accuracy (after pretraining) and the downstream test accuracy (after finetuning). If the attack affects the upstream task accuracy more severely, that means that the downstream task accuracy will suffer too. We tried to design an attack that was going to minimise the negative effect on the upstream test accuracy: poisoning with warmup falls only $\sim 2-3\%$ short of the `clean' upstream test accuracy, which satisfies our first condition. However, this attack too follows the observed linear relationship between upstream and downstream test accuracy, suggesting that isolating an attack to one domain is a harder than we thought. This could be due to the domains being related to each other, meaning that useful feature representations for one domain would be also useful in the other.

It is important to point out that while negative transfer is technically achieved, it is not an accurate measure of unsuccessful knowledge transfer. As pointed out in \citet{wang_characterizing_2019}, negative transfer is a \textit{relative} measure, which depends on multiple factors such as the size of the downstream dataset or the divergence between the two distributions. For example, if the downstream dataset is small, it becomes easier to achieve higher accuracy with transfer learning than training from scratch. Our downstream dataset consists of 5000 labeled training samples, which could be considered relatively big - thus making it a better candidate for negative transfer. On the other hand, there is no significant distribution shift: the downstream labels cover 10 out of the 20 superclasses defined for CIFAR-100 (more details in Section \ref{sec:data}).

In conclusion, while we did achieve negative transfer and high test accuracy on the upstream task, we refrain from declaring our attacks successful for the reasons that we outlined above. If anything, \textbf{the attacks have shown us that the upstream and downstream test accuracy are difficult to attack independently from each other}: in fact, there we observed a linear relationship between them across our attacks. This does not mean that an isolated attack does not exist, merely that it is harder to achieve than we initially thought.

\section{SwAV experiments}
\label{sec:unsup_transfer}
In this section we run an attack on a model which has been pre-trained in a self-supervised fashion with SwAV. The reason why we do experiments on self-supervised models is not just because of their growing popularity, but also because they often serve as a pre-trained base for transfer learning.

\subsection{Baselines}
For the experiments on SwAV, we use a ResNet-50 architecture that was already pretrained on ImageNet. This model achieves 74.3\% top-1 accuracy on ImageNet (this was the reported result\footnote{\url{https://github.com/facebookresearch/vissl/blob/main/MODEL_ZOO.md\#SwAV}} of the model publishers). We further evaluated on the upstream task, where it achieved 77.26\% and on the downstream task where it achieved 96.4\% test accuracy.

\subsection{Attacks}
For the attack, we finetune the entire network on poisoned downstream data for 100 epochs, with checkpoints every 10 epochs. This setting is similar to our `poison after pretraining' attack when we used supervised pretraining. After poisoning, we freeze the ResNet-50 layers and train a linear probe on the learned features. For CIFAR data, we only use the features from the last residual block, while for ImageNet we train separate MLPs for each residual block output, and choose the one with the highest accuracy. The full results can be found in Table \ref{tab:swav}. As we can see, the \textbf{poisoning had a negative effect on all three of the evaluated datasets}: not just the downstream test accuracy. While our goal was to decrease the accuracy of the downstream data only, this result is consistent with what we have experienced in the supervised setting. Another similarity is that \textbf{the effect of poisoning appears early on}: there is little difference between 1, 10 and 100 epochs of poisoning on the CIFAR data. While we did not repeat the same experiment on ImageNet data\footnote{We did not evaluate on ImageNet after 1 and 10 epochs of poisoning due to resource constraints: on a V100 GPU it took around 72 hours to train a linear probe.}, we would expect to see similar results to our CIFAR-100 experiments.

\begin{table}[]
\centering
\begin{tabular}{l|cccc}
\toprule
                 & \multirow{2}{*}{\textbf{before poisoning}} & \multicolumn{3}{c}{\textbf{after poisoning}}                            \\
                 &                 & 1 epoch                  & 10 epochs                                         & 100 epochs \\ \hline
ImageNet         & $74.3\%^{\dagger}$              & -             & -                                                 & 42.66\%    \\
CIFAR upstream   & 77.26\%           &   59.24\%              & 61.91\%                                           & 61.27\%    \\
CIFAR downstream & 96.4 \%  & 88.4\%  & 87.8\% & 87\%      \\
\bottomrule
\end{tabular}
\caption{Top-1 test accuracies of linear probe on learned feature representations before and after poisoning. The base architecture was a ResNet-50 pretrained with SwAV. $^\dagger$: results by model publisher.}
\label{tab:swav}
\end{table}

\section{Summary}
In this chapter, we evaluated the attacks introduced in Section \ref{sec:attacks} in the transfer learning scenario. Our results can be divided into two based on the choice of pretraining: Section \ref{sec:sup_transfer} introduces results on models that were pretrained in a \textit{supervised} manner, while Section \ref{sec:unsup_transfer} shows results in the case of \textit{unsupervised pretraining}.
Our results on the effects of poisoning are consistent across these two settings. \textbf{We observe an overall drop in performance that affects source and target domains equally.}
Additionally, we provide further insights on these attacks in the supervised setting: 

\begin{itemize}
    \item We establish that our attacks indeed lead to bad minima (on the target task): learning curves exhibit unstable behaviour as well as learning less useful feature representations that lead to lower test accuracy.
    \item We observe that one can recover from attacks by training on clean source domain data, while training on target domain data does not help escape the bad minima.
    \item We show a linear relationship between the effect of an attack on target vs. source domain.
\end{itemize}

In conclusion, \textbf{we did not find bad transfer minima, and thus could not prove our original hypothesis} (Hyp. \ref{hyp:first} from Section \ref{sec:methodology}).

{
\renewcommand{\arraystretch}{2.5}
\begin{table}[h]
\centering
\begin{threeparttable}
\begin{tabular}{l|rcc>{\footnotesize}l}
\toprule
\textbf{reference} & \textbf{experiment} & \textbf{us. acc.} & \textbf{ds. acc.} & \textbf{comment} \\ \midrule
    Fig. \ref{fig:scratch_vs_transfer}      &
   \adjustbox{valign=c}{\includegraphics[width=0.1\textwidth]{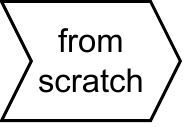}} 
    &      55.53\%       &        74.8\%         &         \\ 
    Fig. \ref{fig:scratch_vs_transfer}      &
   \adjustbox{valign=c}{\includegraphics[width=0.2\textwidth]{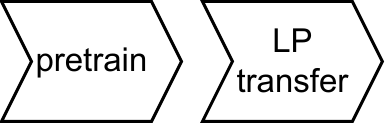}} 
    &      55.53\%       &        \textbf{80.97\% }        &    *     \\ 
    Fig. \ref{fig:scratch_vs_transfer}      &
   \adjustbox{valign=c}{\includegraphics[width=0.1\textwidth]{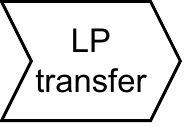}} 
    &      55.53\%       &       45.03\%         &         \\ 
    Fig. \ref{fig:adv_opt_finetune_len}      &
   \adjustbox{valign=c}{\includegraphics[width=0.2\textwidth]{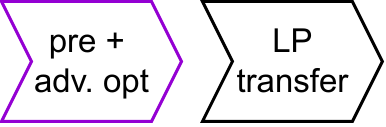}} 
    &       34.07\%        &        58.97\%         &     *    \\ 
    Fig. \ref{fig:pretraining_length}      &      
    \adjustbox{valign=c}{\includegraphics[width=0.3\textwidth]{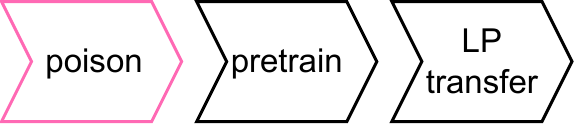}} 
    &       50.04\%       &        79.26\%        &     *   \\
    Fig. \ref{fig:poison_ratio}      &      
    \adjustbox{valign=c}{\includegraphics[width=0.2\textwidth]{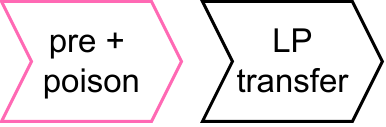}} 
    &       37.1\%       &        54.8\%        &    $p=0.5$  *   \\
    Fig. \ref{fig:poison_len}      &      
    \adjustbox{valign=c}{\includegraphics[width=0.3\textwidth]{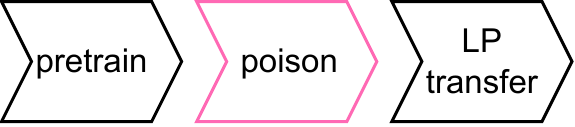}} 
    &       1.15\%       &        41.4\%        &   *   \\
    Fig. \ref{fig:all_attacks}      &      
    \adjustbox{valign=c}{\includegraphics[width=0.3\textwidth]{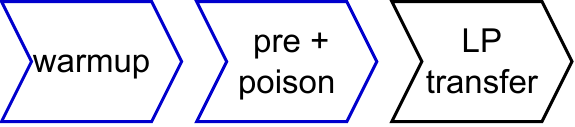}} 
    &       52.94\%       &        72.24\%        &  \multirow{1}{3cm}{Warmup: 20 epochs. **}    \\[3mm] \hline
    Tab. \ref{tab:swav}      &      
    \adjustbox{valign=c}{\includegraphics[width=0.2\textwidth]{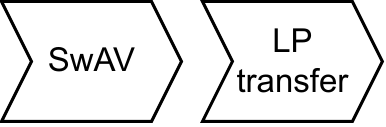}} 
    &       $74.3\%^{\dagger}$     &        \textbf{96.4\%}        &  \multirow{2}{3cm}{Upstream data: ImageNet. Architecture: RN-50. ***}      \\
     Tab. \ref{tab:swav}      &      
    \adjustbox{valign=c}{\includegraphics[width=0.3\textwidth]{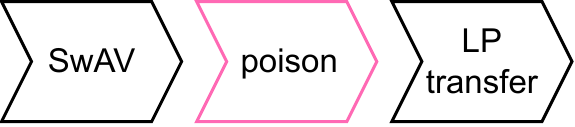}} 
    &       42.66\%       &        87\%        &      \\
          
    \bottomrule
\end{tabular}
\begin{tablenotes}
      \small
      \item $^\dagger$: Not our results.
      \item *: Full transfer results: Appendix \ref{sec:A-full_transfer}.
      \item **: Predecessor of this attack is the l2-anchor attack: Appendix \ref{sec:A-anchor}.
      \item ***: We also experimented with pretraining on CIFAR-100 upstream data: Appendix
      \ref{sec:A-swav}.
    \end{tablenotes}
\end{threeparttable}
\caption{Summary of the experiments that we have done: we report upstream (measured before transfer) and downstream test accuracies (measured after transfer). Top half are on transfer learning w/ supervised pretraining. Upstream/downstream data are partial CIFAR-100; architecture is RN-18, training phases are 100 epochs long each. The results are an average of three runs. Bottom half are pretrained w/ SwAV (unsupervised). Results are from one run.}
\label{tab:all_experiments}
\end{table}
}

%% file: chapters/18Conclusion.tex
\chapter{Conclusion}
\label{chap:conclusion}

\section{Limitations}

The validity of our conclusions rests on three pillars: the definition of bad transfer minima, the choice of attack and the generalizability of our claims. We group the possible limitations of this dissertation according to these three categories.

\paragraph{Bad transfer minima are relative}
As previously discussed, the success of transfer depends on more things than initially apparent: while alignment between source and target distribution is desirable, it is not a necessary criterion\footnote{In the medical image domain transfer learning from models pretrained on ImageNet is common practice, and highly successful (even though the domains are very different) \cite{taher_systematic_2021}.}. Even the way we define similarity between distributions seems to be an important nuance - do we define coarse groups outlining domains (such as medical vs driving images) or do we use fine-grained similarity metrics on the pixel level? Furthermore, \citet{liu_towards_2019}'s  results challenge the common perception that similarity on the input feature space is the most important factor when it comes to positive transfer: they show that similarity between tasks also plays an important role. 
In addition to domain similarity and the choice of task, data size and quality is also instrumental. \citet{wang_characterizing_2019} acknowledge this, and stress the fact that negative transfer is a \textit{relative} measure. Since our definition of bad transfer minima is strongly connected to the definition of negative transfer, we too would like to stress its relative nature.
This makes the interpretation of the results inherently more ambiguous.

\paragraph{Choice of attack}
Another important point to keep in mind is that our choice of attack is not necessarily the best. This applies both to the attack's potential \textit{success}, and its \textit{generality}. It is possible that a different choice of attacks might have been more successful at finding bad transfer minima. Another critique is that our attacks are designed to ruin transfer onto a \textit{specific} downstream task. 
A more general attack  would have allowed us to evaluate it on multiple downstream tasks and make stronger claims (which partially ties into our next point). 

\paragraph{Generalizability of claims}
We evaluated our claims on a limited number of domains and tasks: models were either pretrained on ImageNet or on a subset of CIFAR-100. The only targeted downstream task was a subset of CIFAR-100. While this clearly limits the generalizability of our claims, it was a conscious trade-off between running faster experiments on more attacks and doing large-scale experiments across multiple domains.

\section{Future work}

We see three potential avenues for future work: finding a better definition for bad transfer minima, exploring more (potentially more successful) attacks, and extending our evaluations to more connected scenarios, such as zero-shot learning.

\paragraph{Better definition of bad transfer minima}
As outlined above, our definition of bad transfer minima bears the same limitations as the widely accepted definition of negative transfer. 
Currently it is most commonly used as a binary indicator of success, which does not take into account important nuances that we discussed in the previous section.
Therefore, we propose that future research focus on finding more robust definitions, that potentially correct for some of the most common skewing factors.
For example, common experience dictates that transfer between similar distributions and tasks is easier -  therefore positive transfer between dissimilar domains should be amplified.
Until a definition which is more robust to confounding factors is achieved, we strongly advise future work to judge negative transfer on a case-by-case basis.

\paragraph{Exploring other attacks}
Another possible direction for improvement would be the choice of attack.
Two out of our three presented attacks were based on data manipulation, more precisely by perturbing labels. Exploring attacks that manipulate the feature space  could lend itself to more successful attacks based on the method's success in  targeted poisoning \cite{shafahi_poison_2018,huang_metapoison_2020,geiping_witches_2021}. We also introduced one attack that manipulated the learning process through the choice of loss function, which could be further explored. Gradient matching also seems like a promising direction - achieved either through feature manipulation \cite{geiping_witches_2021} or data reordering attacks \cite{shumailov_manipulating_2021}.

\paragraph{Extending to other scenarios}
It would be most interesting to extend our attacks to few-shot or zero-shot learning scenarios, for example in the natural language processing context.
Large language models have been shown to generalize on a range of tasks using prompt engineering. For example the following sentence "English: I like cake. French: \_\_\_" will likely prompt a language model to translate the text. The extent of achievable tasks and engineering of such prompts has been greatly researched \cite{liu_pre-train_2021}. The advantage of prompt-based learning is that we can achieve few-shot or zero-shot learning, even if we have no or very few labeled data. Future work could try to attack pre-trained language models and see if certain types of tasks (such as translation) can be ``blocked''.

\section{Discussion}
In this dissertation we investigated the connection between generalization on the source task and transferability - in cases where positive transfer is possible. 
Our initial hypothesis was that we could find minima in the parameter space which induce high test error on the target domain (transfer fails) while retaining low test error on the source task.
Motivated by prior work from \citet{zhang_understanding_2017} and \citet{liu_bad_2021}, we use attacks as our means to find minima that do not transfer well.
We give a definition for bad transfer minima, and propose three\footnote{We also tried out a fourth attack, but chose not to include it in the main body because it did not produce the desired results. Results on the l2-anchor attack can be found in Appendix \ref{sec:A-anchor}.} attacks to find it.
Our definition is informal, and relative to a baseline parameter found by a ``clean'' transfer learning algorithm.

Taking these limitations into consideration, we evaluate our proposed attacks on transfer learning (with linear probe\footnote{We also have results on full finetuning, in Appendix \ref{sec:A-full_transfer}}) and self-supervised learning (SwAV), with CIFAR-100 as our dataset which we divide into two parts for simulating source and target data.
While some of our attacks achieve relatively high performance on the source domain \textbf{and} negative transfer on the target domain - which satisfies our definition of bad transfer minima - we refrain from declaring our attack successful.
We suspect that the fulfillment of the second condition of bad transfer minima (negative transfer) was probably more due to circumstance, and is likely to fail on transfer scenarios where the downstream task has less / lower quality data.
While this forced us to be careful in our evaluation of the results, we did arrive at some interesting conclusions.

Our presented attacks are successful at finding bad minima on the target task.
We show that these minima can be reached fast, and that attacks can be scaled.
Our general observation is that while attacks are capable of inducing poor performance on the target task, they also affect the source task test error negatively. 
In fact, we discovered a linear relationship between the test performance of source and target task, which seems to hold across the different attack methods we considered.
Additionally, we found that the effect of dataset poisoning could be reversed by training on clean data from the source domain (after the attack, but before transfer/finetuning). 
To conclude, \textbf{our experiments produced no evidence that could rule out generalization on the source-task being a sufficient condition for transferability.}

While we were not able to prove our initial hypothesis, we presented a well-motivated research direction on the conditions of transferability. 
This work proposed an empirical approach based on adversarial attacks and highlighted some of the most common pitfalls and most promising future directions.
Additionally, our work is the first to propose attacks that target the knowledge transfer between source and target task (to the best of our knowledge).
We hope that the research community finds our conclusions useful and can use our results as a starting point for future directions.

\label{lastcontentpage}

%% file: chapters/21Appendix.tex
\appendix
\chapter{Additional experiments}

\begin{figure}[h]
    \centering
    \includegraphics[width=0.45\textwidth]{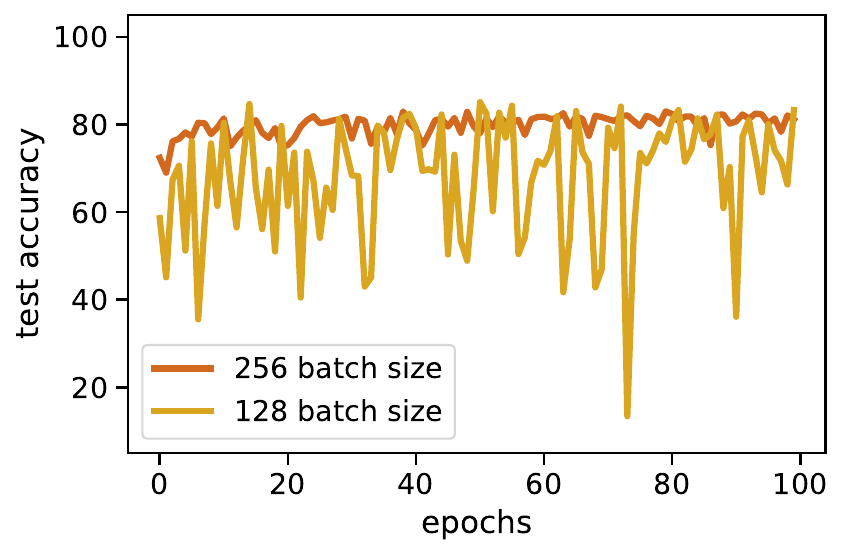}
    \caption{Finetuning learning curve with batch size of 128 vs 256. Evaluated on downstream test data.}
    \label{Afig:128vs256}
\end{figure}

\begin{table}[h]
    \centering
    \begin{tabular}{lc|ccc}
    \toprule
              & \textbf{upstream} & \multicolumn{3}{c}{\textbf{downstream}}              \\
              &          & from scratch    & LP transfer & rand. LP transfer \\ \midrule
            test accuracy & $59.41\% \pm 0.51$ & $60\% \pm 18.08$ & $\textbf{69.68\%} \pm 13.48$ & $21.92\% \pm 6.58$ \\
    \bottomrule
    \end{tabular}
    \caption{128 batch size. Baseline test accuracies on upstream and downstream data. Reported averages and standard deviations are based on 5 runs.}
    \label{Atab:test_acc}
\end{table}

\section{Results with batch size of 128} \label{sec:A-bs_128}

\begin{figure}
    \centering
    \includegraphics[width=0.5\textwidth]{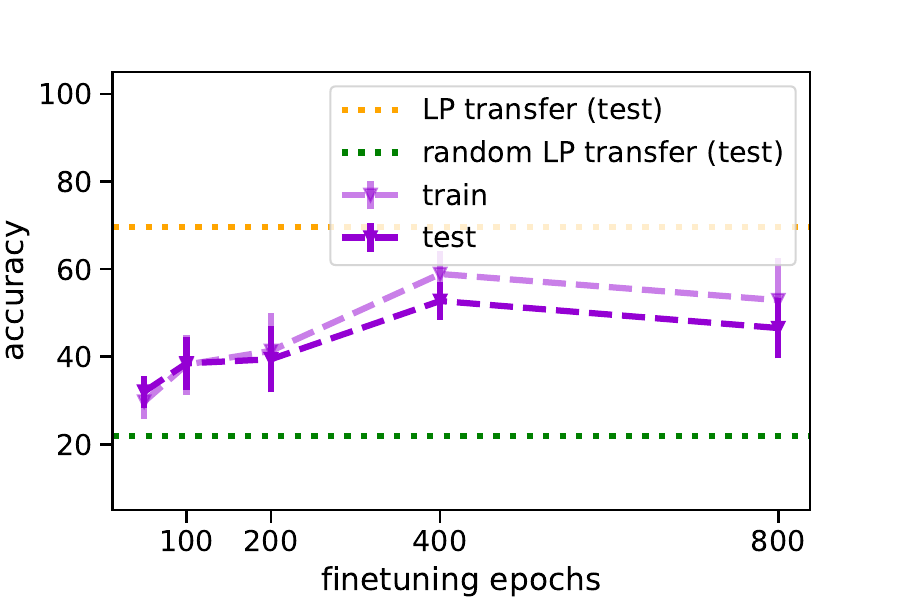}
    \caption{128 batch size. Downstream accuracies on network pretrained with adversarial optimization, then finetuned to downstream task. We compare to the vanilla transfer learning setting (`LP transfer') and to transfer learning on a random initialized network (`random LP transfer'). Reported mean and SE is from 5 runs. 
    }
    \label{Afig:adv_opt_finetune_len}
\end{figure}

\begin{figure}
     \centering
     \begin{subfigure}[b]{0.49\textwidth}
         \centering
         \includegraphics[width=\textwidth]{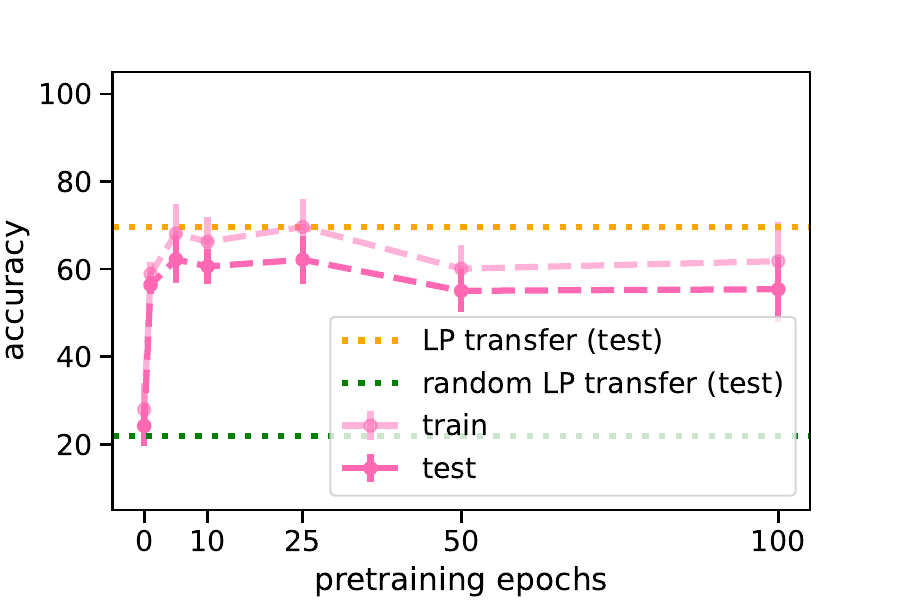}
         \caption{Poisoning before pretraining. Effect of pretraining length.}
         \label{Afig:pretraining_length}
     \end{subfigure}
     \hfill
     \begin{subfigure}[b]{0.49\textwidth}
         \centering
         \includegraphics[width=\textwidth]{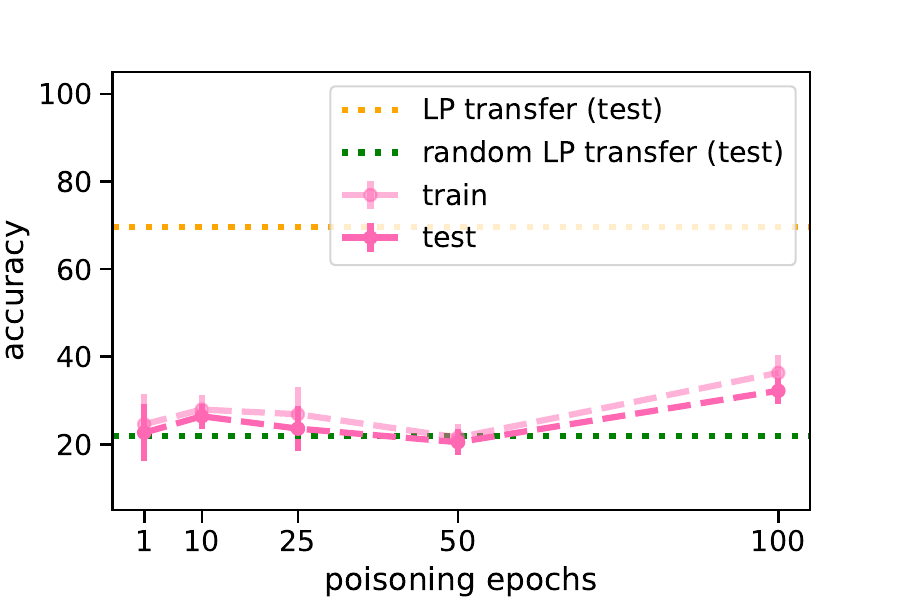}
         \caption{Poisoning after pretraining. Effect of number of poisoning epochs.}
         \label{Afig:poison_len}
     \end{subfigure}
     \hfill
     \begin{subfigure}[b]{0.5\textwidth}
         \centering
         \includegraphics[width=\textwidth]{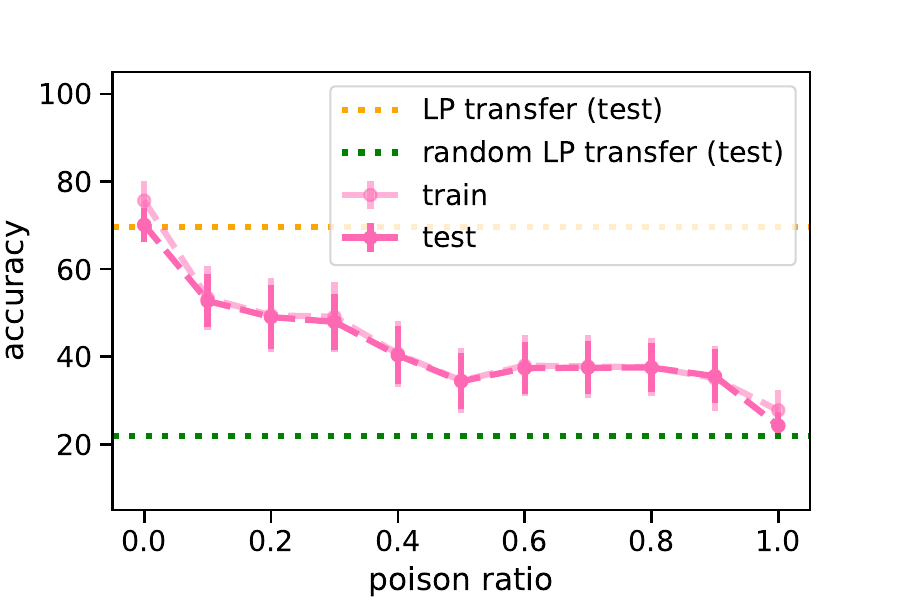}
         \caption{Poisoning during pretraining. Effect of poison ratio $p$.}
         \label{Afig:poison_ratio}
     \end{subfigure}
        \caption{128 batch size. Effect of poisoning attack on finetuned downstream accuracy. We evaluate poisoning \textit{before}, \textit{during} and \textit{after} pretraining. We compare to the vanilla transfer learning setting (`LP transfer') and to transfer learning on a random initialized network (`random LP transfer'). Reported mean and SE is from 5 runs.}
        \label{Afig:three graphs}
\end{figure}

We first ran our experiments with hyperparameters matching those of \citet{liu_bad_2021}, which meant a batch size of 128. Immediately we noticed that our performance on downstream data had an unusually large variance: just looking at the baseline results in Table \ref{Atab:test_acc}, transfer learning with a linear probe had very large standard deviation (this is due to the highly unstable behaviour demonstrated in Figure \ref{Afig:128vs256}). Despite the high noise, our conclusions on the experiments for adversarial optimization and label poisoning remain largely the same. Attacks lead to bad minima relatively fast (Fig. \ref{Afig:poison_len}) and the negative effect of the attack can be varied (Fig. \ref{Afig:poison_ratio}). If we wish to get out of bad minima, it pays off to train on clean upstream data for a few epochs (Fig. \ref{Afig:pretraining_length}) rather than making finetuning longer (Fig. \ref{Afig:adv_opt_finetune_len}).

\newpage
\section{Full transfer results} \label{sec:A-full_transfer}
\begin{figure}
    \centering
    \begin{subfigure}{\textwidth}
        \includegraphics[width=0.49\textwidth]{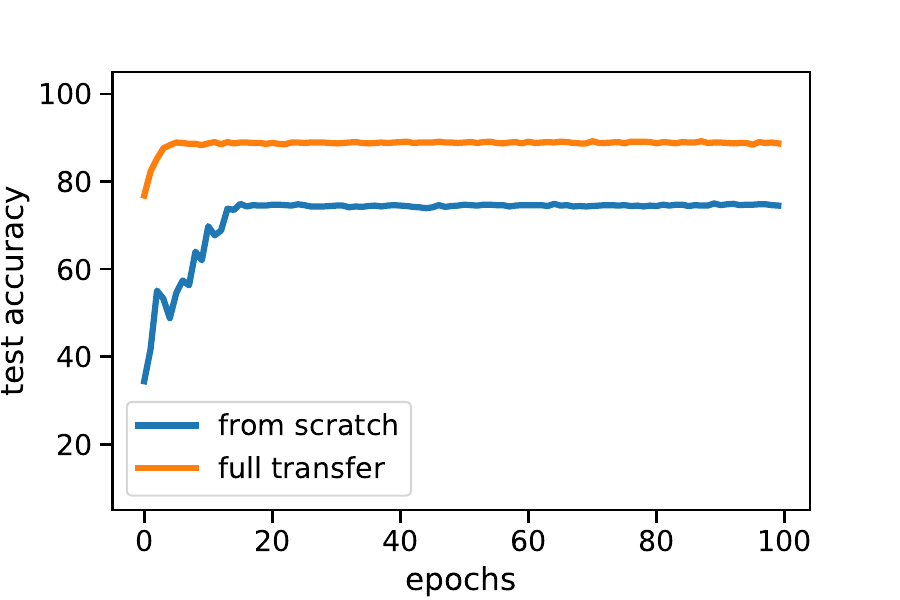}
    \includegraphics[width=0.49\textwidth]{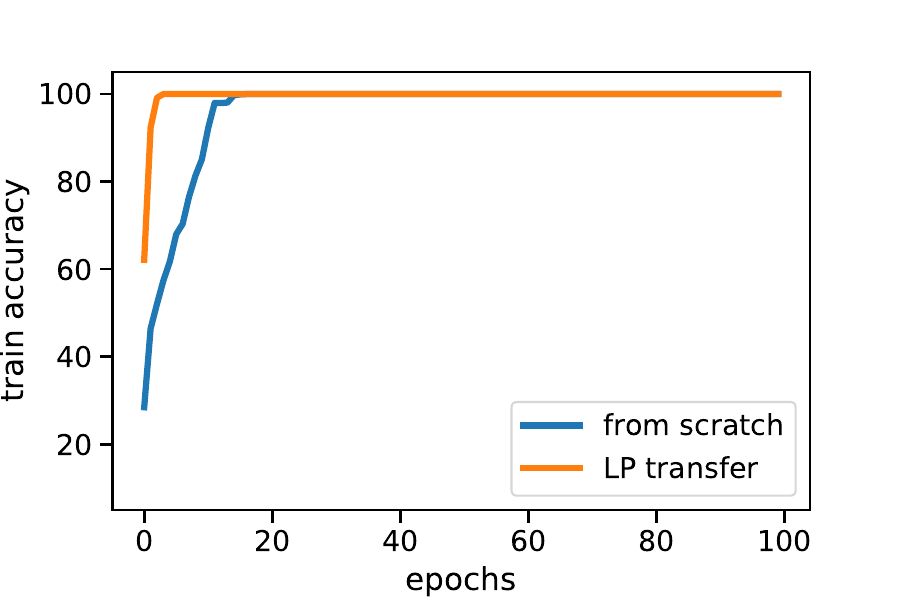}
    \caption{Learning curves of training on downstream data.  Plot is based on one run.}
    \label{Afig:scratch_vs_transfer}
    \end{subfigure}
    
    \bigskip
    
    \begin{subfigure}{\textwidth}
    \centering
        \begin{tabularx}{\linewidth}{lc|cc}
    \toprule
               & \textbf{upstream} & \multicolumn{2}{c}{\textbf{downstream}}              \\
              &          & from scratch    & full transfer \\ \midrule
            test accuracy & $55.53\% \pm 0.62$ & $74.8\% \pm 0.36$ & $\textbf{88.47\%} \pm 1.27$ \\
    \bottomrule
    \end{tabularx}
    \caption{Achieved test accuracies on upstream and downstream data. Reported averages and standard deviations are based on 3 runs.}
    \label{Atab:fulltest_acc}
    \end{subfigure}

    \caption{Full finetuning, 256 batch size.  We compare training from scratch, and transfer learning by fine-tuning the whole network. }
\end{figure}

Another scenario that we tested was the \textit{full transfer} scenario with a batch size of 256. Here, after pretraining, all of the parameters of the model are fine-tuned on the downstream task. This results in an average $88.47\%$ downstream test accuracy, which is a $\sim 14\%$ increase compared to training from scratch (see Table \ref{Atab:fulltest_acc}). 

\begin{figure}
     \centering
     \begin{subfigure}[b]{0.49\textwidth}
         \centering
         \includegraphics[width=\textwidth]{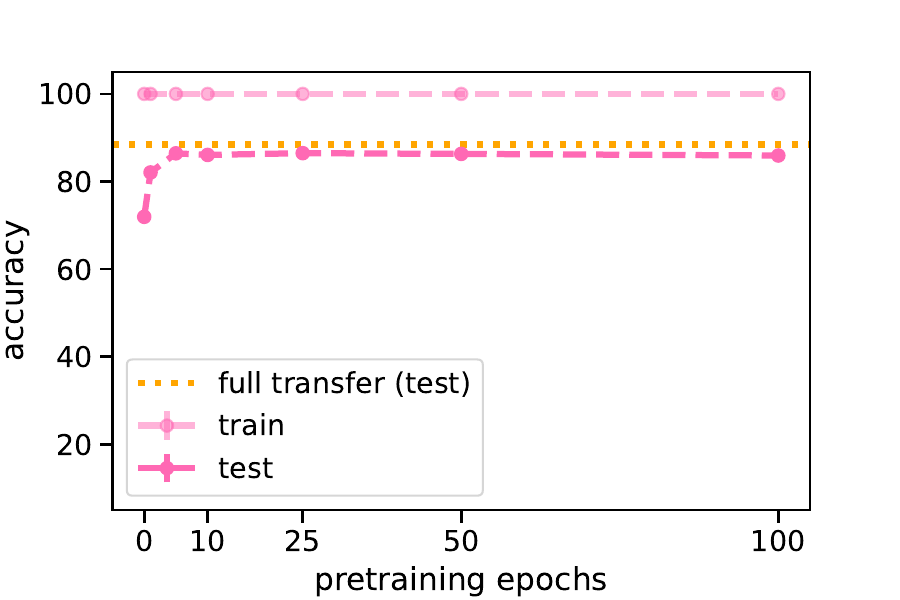}
         \caption{Poisoning before pretraining. Effect of pretraining length.}
         \label{Afig:full_pretraining_length}
     \end{subfigure}
     \hfill
     \begin{subfigure}[b]{0.49\textwidth}
         \centering
         \includegraphics[width=\textwidth]{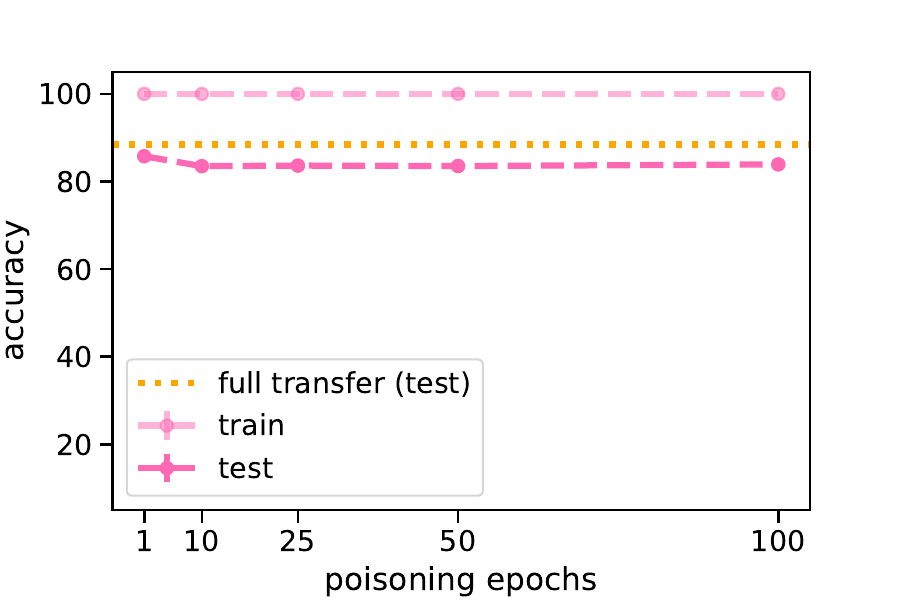}
         \caption{Poisoning after pretraining. Effect of number of poisoning epochs.}
         \label{Afig:full_poison_len}
     \end{subfigure}
     \hfill
     \begin{subfigure}[b]{0.5\textwidth}
         \centering
         \includegraphics[width=\textwidth]{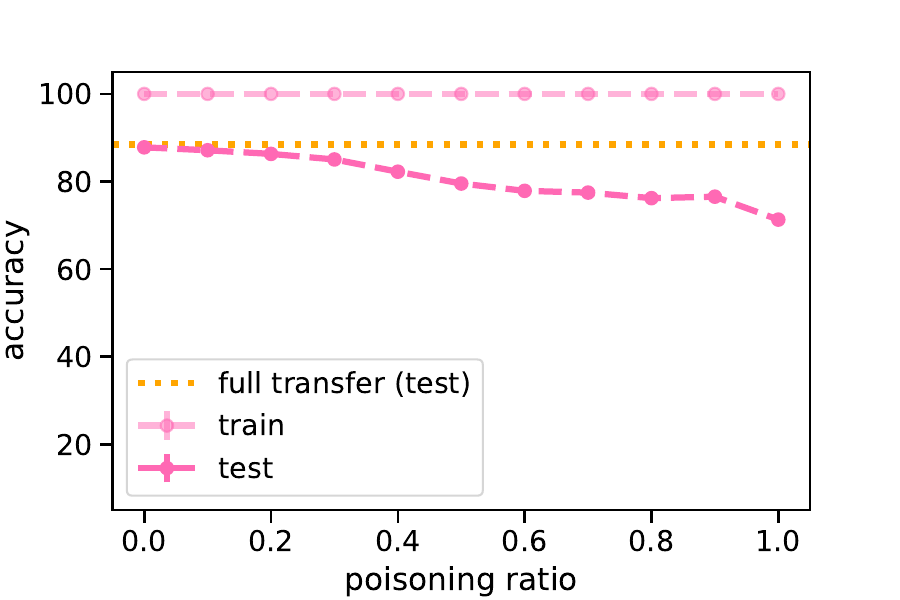}
         \caption{Poisoning during pretraining. Effect of poison ratio $p$.}
         \label{Afig:full_poison_ratio}
     \end{subfigure}
        \caption{Full finetuning, 256 batch size. Effect of poisoning attack on finetuned downstream accuracy. We evaluate poisoning \textit{before}, \textit{during} and \textit{after} pretraining. We compare to the vanilla transfer learning setting (`LP transfer') and to transfer learning on a random initialized network (`random LP transfer'). Reported mean and SE is from 3 runs.}
        \label{Afig:full three graphs}
\end{figure}

If we compare to the linear probe transfer setting from before, here the effects of the attacks are quite subtle. The largest accuracy degradation that we were able to achieve (compared to the full transfer baseline) was in case of `poisoning during pretraining' with $p=1.0$: $71.3\%$ (which is $17.7\%$ lower than the full transfer baseline, see Figure \ref{Afig:full_poison_ratio}). One should also take into consideration that in the $p=1.0$ case, there was no training done at all on the upstream data - therefore a comparison with the training from scratch baseline is more apt (which implies a mere $3.5\%$ degradation). In contrast the attacks on LP transfer were able to reach the naive worst-case baseline (training on random feature representations), which signified a $\sim 30\%$ degradation in accuracy.

\begin{figure}
    \centering
    \includegraphics[width=0.5\textwidth]{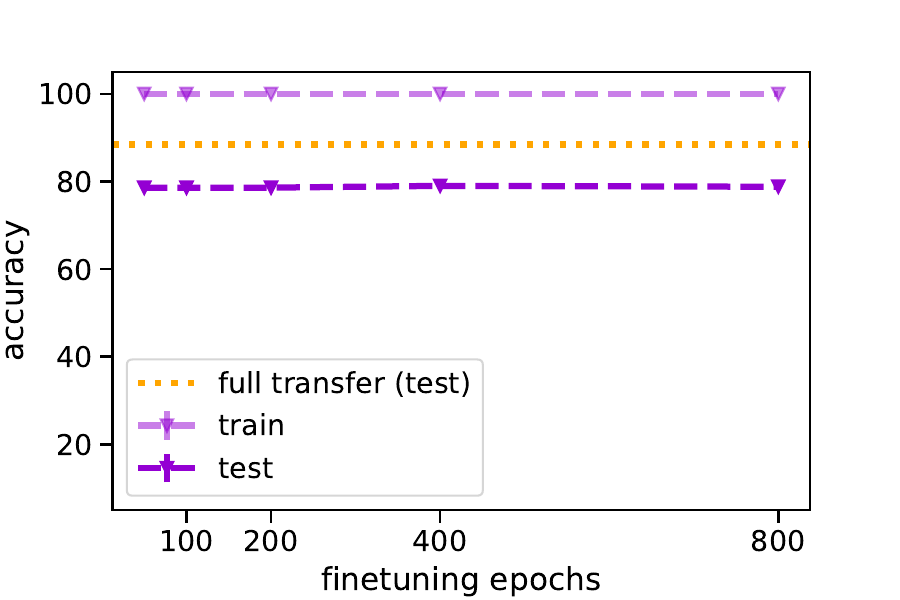}
    \caption{Full finetuning, 256 batch size. Downstream accuracies on network pretrained with adversarial optimization, then finetuned to downstream task. We compare to the vanilla transfer learning setting (`LP transfer') and to transfer learning on a random initialized network (`random LP transfer'). Reported mean and SE is from 3 runs. 
    }
    \label{Afig:full_dv_opt_finetune_len}
\end{figure}

Our views on clean training after the attack remain the same: while increasing the number of epochs of the finetuning phase makes no difference (Fig. \ref{Afig:full_dv_opt_finetune_len}), a few epochs of pretraining can fully recover the downstream test accuracy (Fig. \ref{Afig:full_pretraining_length}). 
Interestingly enough, pretraining \textit{before} an attack also seems to make a difference. Instead of $\sim 17\%$, we only get $\sim 5\%$ lower accuracy if we compare $p=1.0$ from Fig. \ref{Afig:full_poison_ratio} and Fig. \ref{Afig:full_poison_len}.

\newpage
\section{$l2$-anchor attack} \label{sec:A-anchor}

The main motivation behind this attack was to reduce the effect of the attack on upstream test accuracy as much as possible. Inspired by the EWC loss from \citet{kirkpatrick_overcoming_2017} - which is used to prevent catastrophic forgetting - we also wish to encourage retaining useful representations for the upstream task, while learning harmful feature representations for the downstream task. Assuming that the bad transfer minimum is in the same basin as the one that can be reached with clean training, we can define a loss function which penalizes getting further away from it.

\begin{figure}
    \begin{subfigure}[b]{0.45\textwidth}
    \centering
        \includegraphics[width=\textwidth]{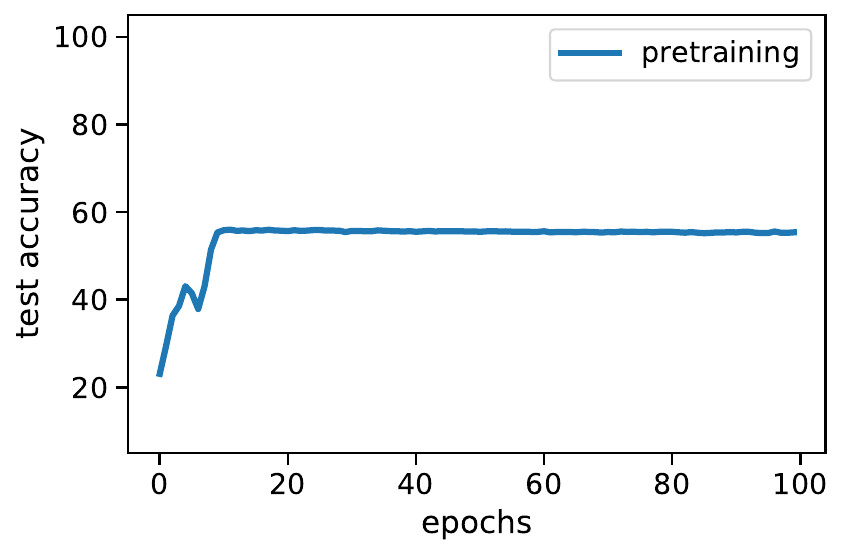}
        \caption{Pretraining learning curve (no attack).}
        \label{Afig:pre_learning_curve}
    \end{subfigure}
    \hfill
    \begin{subfigure}[b]{0.53\textwidth}
    \centering
        \includegraphics[width=\textwidth]{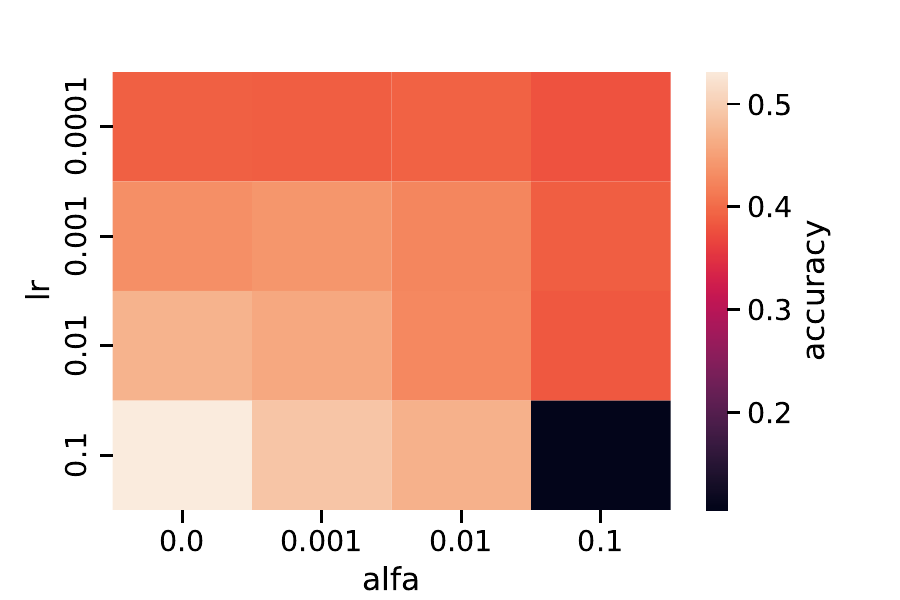}
        \caption{After l2-anchor attack.}
        \label{Afig:l2_anchor}
    \end{subfigure}
    
    \caption{Upstream test accuracies. Training on upstream data converges after $\sim 20$ epochs. We test different values of alfa and learning rate during l2-anchor attack (with a warmup phase of 20 epochs). Reported accuracies are from one run.}
    
\end{figure}

Looking at the learning curve of clean training (Figure \ref{Afig:pre_learning_curve}), we can see that the model converges after 20 epochs. Thus we can use the minimum that we reach after 20 epochs as an \textit{anchor} while we train on poisoned data (poison \textit{during} pretraining). Thus, we can define the following loss function, where $\theta_a$ is the anchor and $\theta$ is the current weight parameters that we optimize on and $\mathcal{L}_{CE}$ is the cross-entropy loss:

\begin{equation}
    \mathcal{L} = \mathcal{L}_{CE} + \alpha || \theta_a - \theta ||_2
\end{equation}

, where $\alpha$ is a parameter controlling the penalty term, which is the $l2$ distance between the anchor and our current parameters. We experiment with different values of alpha and learning rate, which we present in Figure \ref{Afig:l2_anchor}. Unfortunately, the penalty term results in lower test accuracy in all cases. However the warmup phase does seem to be beneficial, leading to upstream test accuracy close to our clean training baseline.

\newpage
\section{Training SwAV on CIFAR-100} \label{sec:A-swav}

We train CIFAR-100 on a ResNet-18 architecture with SwAV, with hyperparameters from a repository \footnote{\url{https://github.com/abhinavagarwalla/swav-cifar10}} where one of the contributors is one of the original authors. They performed hyperparameter tuning for training CIFAR-10 on a ResNet-50 architecture. This of course does not guarantee that the hyperparameters will be optimal for our setting, but due to resource constraints\footnote{Training for 500 epochs took approximately 16 hours on a single V100 GPU.} we did not perform a hyperparameter sweep. For training, we used the Vissl framework \footnote{Vissl: \url{https://github.com/facebookresearch/vissl}} for self-supervised learning research. We extended it to support the ResNet-18 architecture. We used no multicrop, a batch size of 512, set the number of prototypes $K=30$, and initialized the SwAV loss function with temperature $\tau=0.5$, 3 iterations for the Sinkhorn-Knopp \cite{cuturi_sinkhorn_2013} algorithm (for calculating $Q$), $\epsilon=0.05$ and no memory bank (queue) for storing past feature representations. Training was performed for 500 epochs with SGD as optimizer, with a momentum of $0.9$ and a cosine learning rate scheduler with base set to $0.06$ and end learning rate set to $0.0006$.

We trained a linear probe on the learned feature representations (on the output of the last residual block) and evaluated it on the CIFAR-100 test set. Unfortunately, the highest top-1 accuracy that we could achieve was 32.93\%, which is well below the supervised baseline on the same architecture. Because of these dissatisfactory baseline results, we decided not to pursue further experiments involving attacks. We achieved this result with a batch size of 512, and SGD with cosine parameter scheduler with start value of 0.03 and momentum of 0.9. Again, we used the same hyperparameters as the already mentioned repository.